\documentclass[runningheads]{llncs}

\usepackage{eccv}

\usepackage{eccvabbrv}

\usepackage{graphicx}
\usepackage{booktabs}

\usepackage[accsupp]{axessibility}  

\usepackage{url}

\usepackage{tikz}
\usepackage{comment}
\usepackage{enumitem}

\usepackage{color}
\usepackage{verbatim}
\usepackage{multirow}
\usepackage{enumitem}
\usepackage{mathrsfs}
\usepackage{bm}
\usepackage[accsupp]{axessibility}  

\usepackage{graphicx}
\usepackage{subcaption}
\usepackage{pifont}

\usepackage[pagebackref,breaklinks,colorlinks,citecolor=eccvblue]{hyperref}
\usepackage{hyperref}

\usepackage{orcidlink}
\newcommand{\cmark}{\ding{52}}
\newcommand{\xmark}{\ding{56}}

\usepackage{colortbl} 

\definecolor{mygreen}{RGB}{34,139,34}
\definecolor{mygray}{gray}{0.6}
\definecolor{mygray-bg}{gray}{0.9}

\begin{document}

\title{DECap: Towards Generalized Explicit Caption Editing via Diffusion Mechanism} 

\titlerunning{Towards Generalized Explicit Caption Editing via Diffusion Mechanism}

\author{
    Zhen Wang\inst{1}
    \and 
    Xinyun Jiang\inst{2} \and 
    Jun Xiao\inst{1} \and 
    Tao Chen\inst{3} \and 
    Long Chen\inst{4}\thanks{Corresponding author.}
}

\authorrunning{Z. Wang and L. Chen et al.}
%
\institute{$^1$Zhejiang University \quad
$^2$Massachusetts Institute of Technology \quad \\
$^3$Tongdun Technology\quad
$^4$The Hong Kong University of Science and Technology\\
\email{zju\_wangzhen@zju.edu.cn, junx@cs.zju.edu.cn, longchen@ust.hk}
}
\maketitle

\begin{abstract}
Explicit Caption Editing (ECE) --- refining reference image captions through a sequence of explicit edit operations (\eg, \texttt{KEEP}, \texttt{DETELE}) --- has raised significant attention due to its explainable and human-like nature. After training with carefully designed reference and ground-truth caption pairs, state-of-the-art ECE models exhibit limited generalization ability beyond the original training data distribution, \ie, they are tailored to refine content details only in in-domain samples but fail to correct errors in out-of-domain samples. To this end, we propose a new Diffusion-based Explicit Caption editing method: \textbf{DECap}. Specifically, we reformulate the ECE task as a denoising process under the diffusion mechanism, and introduce innovative edit-based noising and denoising processes. Thanks to this design, the noising process can help to eliminate the need for meticulous paired data selection by directly introducing word-level noises for training, learning diverse distribution over input reference caption. The denoising process involves the explicit predictions of edit operations and corresponding content words, refining reference captions through iterative step-wise editing. To further efficiently implement our diffusion process and improve the inference speed, DECap discards the prevalent multi-stage design and directly generates edit operations and content words simultaneously. Extensive ablations have demonstrated the strong generalization ability of DECap in various scenarios. More interestingly, it even shows great potential in improving the quality and controllability of caption generation.
  \keywords{Explicit Caption Editing \and Image Captioning \and Diffusion Model \and Out-of-Distribution (OOD) Evaluation}
\end{abstract}

\section{Introduction}
\label{sec:intro}

\begin{figure*}[t]
  \centering
  \includegraphics[width=0.99\linewidth]{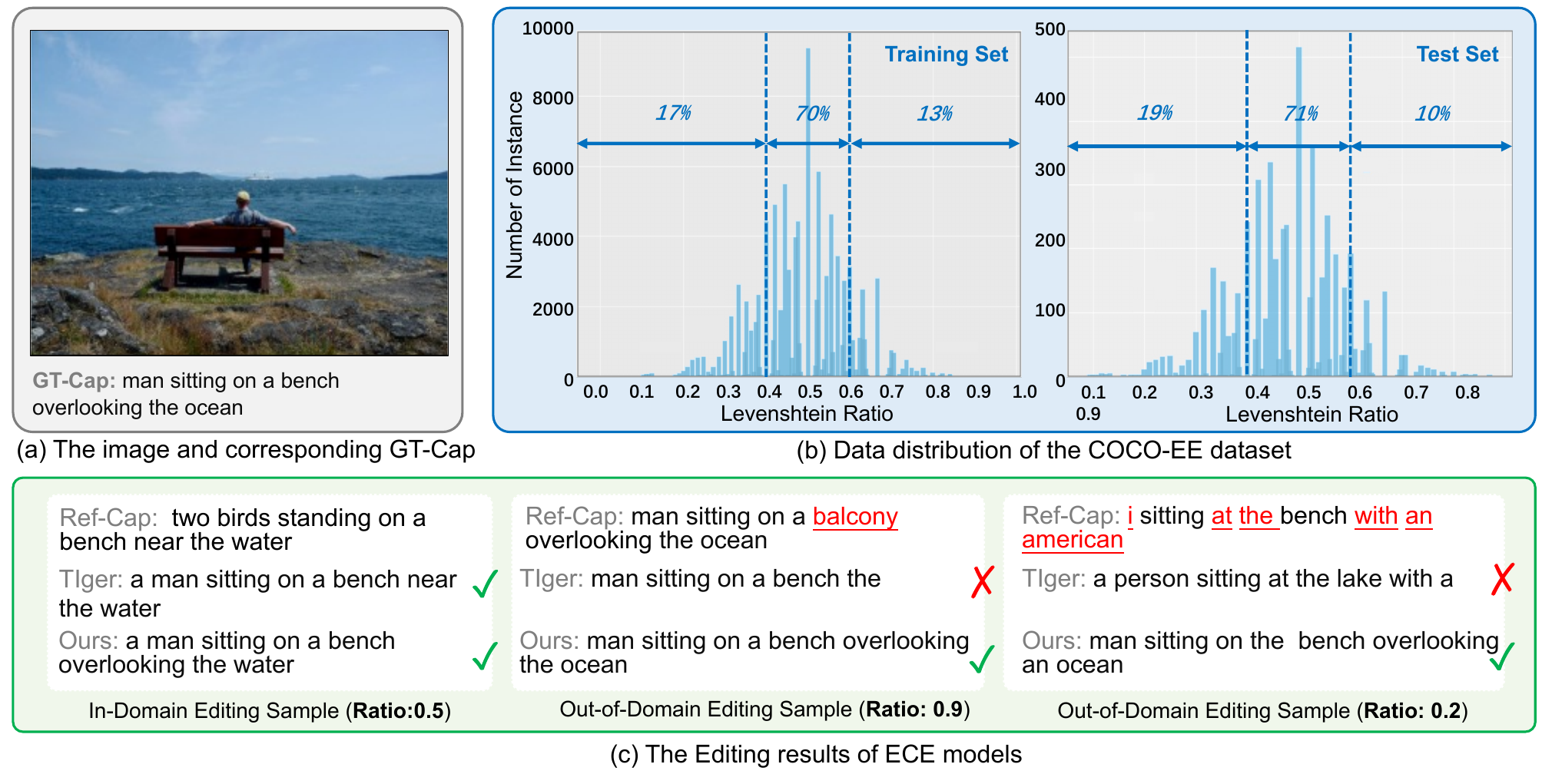}
  \vspace{-1em}
  \caption{\textbf{(a)} An image example and its corresponding ground-truth caption (GT-Cap). \textbf{(b)} Data distribution of the COCO-EE dataset~\cite{wang2022explicit}. The distribution of the training set and test set are very similar, where most of the editing instances have ratios ranging from 0.4 to 0.6. \textbf{(c)} Editing results of state-of-the-art ECE model TIger and our DECap. The \emph{in-domain} Ref-Cap sample is from the COCO-EE test set, and \emph{out-of-domain} Ref-Cap samples are constructed by replacing the GT-Cap with other words, \eg, predicted by BERT (or sentences generated by pretrained captioning models).}
 \label{fig:intro}
\end{figure*}

Explicit Caption Editing (ECE), emerging as a novel task within the broader domain of caption generation, has raised increasing attention from the multimodal learning community~\cite{wang2022explicit}. As shown in Fig.~\ref{fig:intro}(a)(c), given an image and a reference caption (Ref-Cap), ECE aims to explicitly predict a sequence of edit operations, which can translate the Ref-Cap to ground-truth caption (GT-Cap). Compared to conventional image captioning methods which generate captions from scratch ~\cite{vinyals2015show,xu2015show,chen2017sca,anderson2018bottom}, ECE aims to enhance the quality of existing captions in a more explainable, efficient, and human-like manner.

Currently, existing ECE methods primarily rely on two prevalent benchmarks for model training and evaluation, \ie, COCO-EE and Flickr30K-EE~\cite{wang2022explicit}. Specifically, both datasets are carefully constructed to emphasize the refinement of content details while preserving the original caption structure. As shown in Fig.~\ref{fig:intro}, each ECE instance consists of an image along with a Ref-Cap (\eg, \texttt{two birds standing on a bench near the water}) and a corresponding GT-Cap (\texttt{man sitting on a bench overlooking the ocean}). For this in-domain sample, state-of-the-art ECE models can effectively improve the quality of the Ref-Cap. By ``in-domain'', we mean that the test set of existing ECE benchmarks has a similar distribution with its training set (Fig.~\ref{fig:intro}(b))\footnote{In this paper, we use the Levenshtein ratio (\texttt{ratio}) to quantify the similarity between two captions by considering their lengths and edit distances. The range of \texttt{ratio} is from 0 to 1, where a higher value indicates higher similarity.\label{ft:ratio}}. However, we found that existing ECE models have limited generalization ability when faced with out-of-domain samples. Take the model TIger~\cite{wang2022explicit} as an example, given a highly similar Ref-Cap with a single wrong word (\texttt{man sitting on a \underline{balcony} overlooking the ocean}), although it corrected the wrong word, it also removes other accurate words. Meanwhile, when faced with more irrelevant Ref-Caps (\eg, \texttt{\underline{i} sitting \underline{at} \underline{the} bench \underline{with} \underline{an} \underline{american}}), TIger even fails to correct all errors or introduce sufficient accurate details. Obviously, this limited generalization ability will limit their utilization in real-world scenarios, as we hope our ECE models can help to edit or refine different sentences.

To address this limitation, we propose a novel diffusion-based ECE model, denoted as \textbf{DECap}, which reformulates the ECE task as a series of deonising process steps. Specifically, we design an edit-based noising process that constructs editing samples by introducing word-level noises (\ie, random words) directly into the GT-Caps to obtain Ref-Caps. This noising process is parameterized by the distributions over both edit operations (\eg, \texttt{KEEP}, \texttt{DELETE}, \texttt{INSERT}, and \texttt{REPLACE}) and caption lengths, which can not only avoid the meticulous selection of Ref-GT caption pairs but also help ECE models to learn a more adaptable distribution over Ref-Caps, capturing a broader spectrum of editing scenarios. Then, we train model DECap to refine Ref-Caps through an edit-based denoising process, which contains the iterative predictions of edit operations and content words. Meanwhile, DECap discards the prevalent multi-stage architecture designs and directly generates edit operations and content words simultaneously, which can significantly accelerate the inference speed with simple Transformer encoder architectures. Extensive ablations have demonstrated that DECap can not only achieve outstanding editing performance on the challenging ECE benchmarks but can also further enhance the quality of model-generated captions. Meanwhile, it even achieves competitive caption generation performance with existing diffusion-based image captioning models. Furthermore, DECap even shows potential for word-level controllable captioning, which is beyond the ability of existing controllable captioning models~\cite{cornia2019show,deng2020length,chen2021human}. In summary, \emph{DECap realizes a strong generalization ability across various in-domain and out-of-domain editing scenarios, and showcases great potential in improving the quality and controllability of caption generation, keeping the strong explainable ability. \textbf{With such abilities, our DECap can serve as an innovative and uniform framework that can achieve both caption editing and generation}}.

In summary, we make several contributions in this paper: 1) To the best of our knowledge, we are the first work to point out the poor generalization issues of existing ECE models, and propose a series of caption editing scenarios for generalization ability evaluation. 2) DECap is the first diffusion-based ECE model, which pioneers the use of the discrete diffusion mechanism for ECE. 3) DECap shows strong generalization ability across various editing scenarios, achieving outstanding performance. 4) DECap has a much faster inference speed than existing ECE methods.

\section{Related Work}

\noindent\textbf{Explicit Caption Editing (ECE).} Given the image, ECE aims to refine existing Ref-Caps through a sequence of edit operations, which was first proposed by Wang~\etal~\cite{wang2022explicit}. Specifically, by realizing refinement under the explicit traceable editing path composed of different edit operations, this task encourages models to enhance the caption quality in a more explainable and efficient manner. However, existing ECE benchmarks are carefully designed, targeting on the refinement of specific content details, which leads to a limited model generalization ability across diverse real-world editing scenarios beyond the training data distribution. Meanwhile, existing editing models~\cite{mallinson2020felix,wang2022explicit,reid2022learning} tend to perform editing with multiple sub-modules sequentially. For example, conducting the insertion operation by first predicting the \texttt{ADD} operation, then applying another module to predict the specific word that needs to be added. In this paper, we construct Ref-Caps by directly noising the GT-Caps at word-level through a novel edit-based noising process, allowing the model to capture various editing scenarios during training. We further optimize model architecture to predict both edit operations and content words parallelly, which can significantly accelerate the editing speed.

\noindent\textbf{Diffusion-based Captioning Models.} Taking inspiration from the remarkable achievements of diffusion models in image generation~\cite{austin2021structured,rombach2022high}, several pioneering works have applied the diffusion mechanism for caption generation. Existing diffusion-based captioning works can be categorized into two types: 1) \textbf{Continuous Diffusion}: They aim to convert discrete words into continuous vectors (\eg, word embeddings~\cite{he2023diffcap} and binary bits~\cite{chen2022analog,luo2023semantic}) and apply the diffusion process with Gaussian noises. 2) \textbf{Discrete Diffusion}: They aim to extend the diffusion process to discrete state spaces by directly noising and denoising sentences at the token level, such as gradually replacing tokens in the caption with a specific [\texttt{MASK}] token and treating the denoising process as a multi-step mask prediction task starting from an all [\texttt{MASK}] sequence~\cite{zhu2022exploring}. As the first diffusion-based ECE model, in contrast to iterative mask replacement, which only trains the ability to predict texts for [\texttt{MASK}] tokens, our edit-based noising and denoising process can help our model to learn a more flexible way of editing (\eg, insertion, deletion, and replacement) by different edit operations. Meanwhile, our model shows its great potential in directly editing random word sequences, which achieves competitive performance to diffusion-based captioning models.

\section{Diffusion-based Explicit Caption Editing}

In this section, we first give a brief introduction of the task ECE and the preliminaries about discrete diffusion mechanism in Sec.~\ref{method:formulation}. Then, we show the edit-based noising and denoising process in Sec.~\ref{method:discrete_diffusion}. We introduce our model architecture in Sec.~\ref{method:model}. Lastly, we demonstrate the details of training objectives and inference process in Sec.~\ref{method:train_infer}.

\subsection{Task Formulation and Preliminaries}
\label{method:formulation}

\textbf{Explicit Caption Editing.} Given an image $I$ and a reference caption (Ref-Cap) $\bm{x}^r=\{w_r^1, ..., w_r^n\}$ with $n$ words, ECE aims to predict a sequence of $m$ edit operations $E=\{e^1, ..., e^m\}$ to translate the Ref-Cap close to the ground-truth caption (GT-Cap) $\bm{x}_0=\{w^1_0, ..., w^k_0\}$ with $k$ words.

\noindent\textbf{Edit Operations.} Normally, different ECE models may utilize different edit operations. While early models mainly focus on the reservation (\eg, \texttt{KEEP}) and deletion (\eg, \texttt{DELETE}) of existing contents, and the insertion (\eg, \texttt{ADD}, \texttt{INSERT}) of new contents, subsequent works~\cite{mallinson2020felix,reid2022learning} have demonstrated that incorporating replacement can improve editing performance more efficiently. Acknowledging this established insight and without loss of generality, in this paper, we utilize the four Levenshtein edit operations\footnote{We are open to investigating other operations (\eg, \texttt{REORDER}) in the future.} for both the noising and denoising process, including: 1) \texttt{KEEP}, the keep operation preserves the current word unchanged; 2) \texttt{DELETE}, the deletion operation removes the current word; 3) \texttt{INSERT}, the insertion operation adds a new word after the current word; 4) \texttt{REPLACE}, the replacement operation overwrites the current word with a new word.

\noindent\textbf{Discrete Diffusion Mechanism.} 
For diffusion models in the discrete state spaces for text generation, each word of sentence $\bm{x}_t$ is a discrete random variable with $K$ categories, where $K$ is the word vocabulary size. Denoting $\bm{x}_t$ as a stack of one-hot vectors, the noising process is written as:
\begin{equation}
\small
    q({\bm{x}}_t | {\bm{x}}_{t-1})={Cat}(\bm{x}_{t};{p}=\bm{x}_{t-1}\bm{Q}_t),
\end{equation}
where ${Cat}(\cdot)$ is a categorical distribution and $\bm{Q}_t$ is a transition matrix applied to each word in the sentence independently: $[\bm{Q}_t]_{i,j}=q(w_t=j | w_{t-1}=i)$. Existing discrete diffusion text generation works~\cite{austin2021structured,he2022diffusionbert,zhu2022exploring} mainly follow the noising strategy of BERT~\cite{devlin2018bert}, where each word stays unchanged or has some probability transitions to the \texttt{[MASK]} token or other random words from the vocabulary. Meanwhile, they incorporate an absorbing state for their diffusion model as the \texttt{[MASK]} token: 
\begin{align}
\small
    [\bm{Q}_t]_{i,j} = 
    \begin{cases}
        1 & \text{if}\  i = j = \texttt{[MASK]}, \\
        \beta_t & \text{if}\  j = \texttt{[MASK]}, i \ne \texttt{[MASK]}, \\
        1 - \beta_t & \text{if}\  i = j \ne \texttt{[MASK]}.
    \end{cases}
\end{align}

After a sufficient number of noising steps, this Markov process converges to a stationary distribution $q(\bm{x}_T)$ where all words are replaced by the \texttt{[MASK]} token. Discrete diffusion works then train their models to predict target words for \texttt{[MASK]} tokens as the denoise process $p_\theta({\bm{x}_{t-1} | \bm{x}_t}, t)$, and generate the sentence by performing a series of denoising steps from an all \texttt{[MASK]} token sequence:
\begin{equation}
\small
P_\theta(\bm{x}_{0})=\textstyle{\prod}_{t=1}^Tp_\theta(\bm{x}_{t-1} | \bm{x}_t, t).
\end{equation}

\subsection{Discrete Diffusion for ECE}
\label{method:discrete_diffusion}

Taking inspiration from the discrete process where a noised sentence is iteratively refined into the target,  we reformulate ECE training with a discrete diffusion mechanism and parameterize the noising and denoising process by way of sampled discrete edit operations applied over the caption words. \emph{This noising and denoising process can clearly mitigate the need for paired Ref-GT caption pairs, as we only need to conduct the diffusion process on original GT-Caps}.

\begin{figure*}[t]
  \centering
  \includegraphics[width=0.99\linewidth]{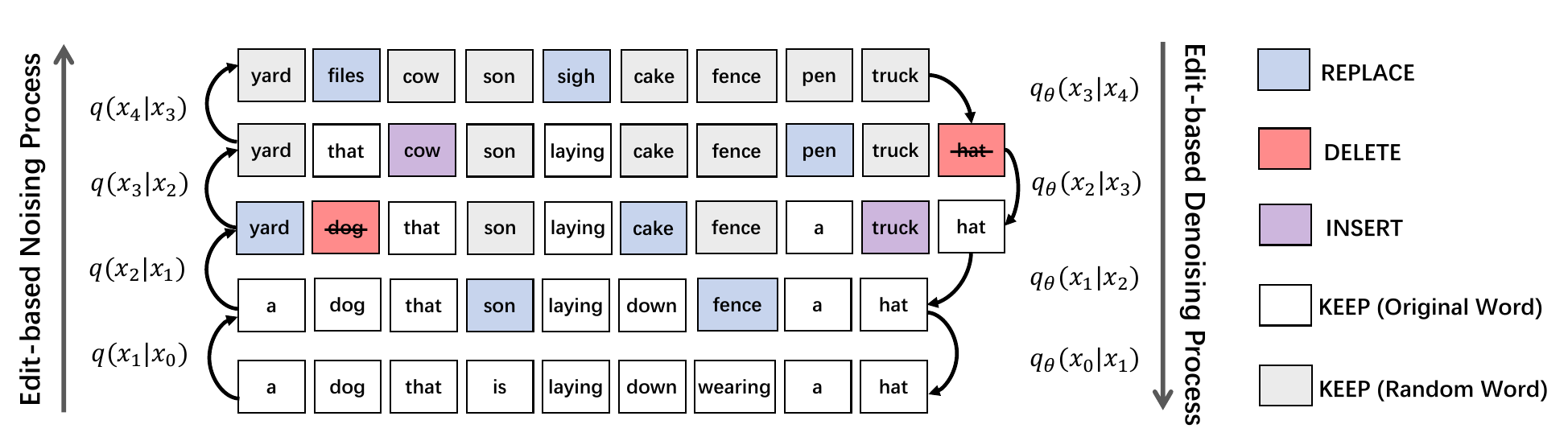}
  \vspace{-1em}
  \caption{Edit-based noising process for DECap. \textcolor[RGB]{143,170,220}{\textbf{Blue}} represents the \texttt{REPLACE} operation, \textcolor[RGB]{245,80,55}{\textbf{red}} represents the \texttt{DELETE} operation, \textcolor[RGB]{164,149,179}{\textbf{purple}} represents the \texttt{INSERT} operation, \textbf{white} and \textcolor[RGB]{150,150,150}{\textbf{grey}} represent the \texttt{KEEP} operation for original word and random word respectively.}  
  \label{fig:noising_process}
\end{figure*}

\noindent\textbf{Edit-based Noising Process.}
Different from directly transiting one word to another, the edit-based noising process gradually adds word-level noises to the caption $\bm{x}_{t-1}$ based on different edit operations. For any time step $t\in (0,T]$, the edit-based noising process is defined as
\begin{equation}
\small
    q(\bm{x}_t| \bm{x}_{t-1}) = p(\bm{x}_t|\bm{x}_{t-1},E^N_t) \cdot {Cat}(E^N_t;{p}=\bm{x}_{t-1}\bm{Q}_t),
\end{equation}
where ${Cat}(\cdot)$ is a categorical distribution and $\bm{Q}_t$ here is a transition matrix assigning edit operation for each word in the caption $\bm{x}_{t-1}$ independently: $[\bm{Q}_t]_{i,j}=q(e_t=j | w_{t-1}=i)$. Subsequently, $E^N_t=\{e^1_t, e^2_t, ..., e^l_t\}$ is a sequence of noising edit operations which has the same length with the caption $\bm{x}_{t-1}=\{w^{1}_{t-1}, w^{2}_{t-1}, ..., w^{l}_{t-1}\}$\footnote{Generally, captions' length may vary in different steps. For simplicity, we slightly use $l$ to denote the length of all other $\bm{x}_t$ captions in this paper.}, where each edit operation $e^{i}_{t}$ is operated on the corresponding word $w^{i}_{t-1}$ to get $\bm{x}_t$. Specifically, $\bm{Q}_t$ is parameterized by the distribution over both edit types and GT-Cap length $k$ with an absorbing state as the random word (\text{RW}). 
\begin{align}
\footnotesize
    [\bm{Q}_t]_{i,j} \!=\! 
    \begin{cases}
        1 & \text{if}\  j = \texttt{KEEP},~~~~~~~ i= \text{RW}, \\
        \alpha^k_t & \text{if}\  j = \texttt{REPLACE}, i \ne \text{RW},\\
        \beta^k_t & \text{if}\  j = \texttt{DELETE},~~ i \ne \text{RW}, \\
        \gamma^k_t & \text{if}\  j = \texttt{INSERT},~~ i \ne \text{RW}, \\
       1-\alpha^k_t-\beta^k_t-\gamma^k_t, & \text{if}\  j = \texttt{KEEP},~~~~~~~ i \ne \text{RW}.
    \end{cases}
\end{align}
Subsequently, as the example shown in Fig.~\ref{fig:noising_process}, being operated with $e_{t}$, each word $w_{t-1}$ has a probability of $\alpha^k_t$ to be replaced by another random word, has a probability of $\beta^k_t$ to be removed from the caption, and has a probability of $\gamma^k_t$ to be added with a random word after it, leaving the probability of $\delta^k_t =  1-\alpha^k_t-\beta^k_t-\gamma^k_t$ to be unchanged. Accordingly, the distribution over the GT-Cap length $k$ can ensure a smooth increase of noised words for each noising step from $\bm{x}_0$ to $\bm{x}_T$.

The distribution over edit types ensures the balance between different noising operations and the learning of different denoising abilities: 1) To learn the ability to \texttt{INSERT} new words, we remove words by \texttt{DELETE} operation. 2) To learn the ability to \texttt{DELETE} incorrect words, we add random words by \texttt{INSERT} operation. 3) To learn the ability to directly \texttt{REPLACE} incorrect words, we change current words into random words by \texttt{REPLACE} operation. 4) To learn the ability to \texttt{KEEP} correct content, we leave the correct words unchanged by \texttt{KEEP} operation. Meanwhile, if the word has already been noised into the random word, it will not be re-noised again. Through sufficient noising steps $T$, the caption will be noised into a random word sequence.

\begin{figure*}[t]
  \centering
  \includegraphics[width=0.99\linewidth]{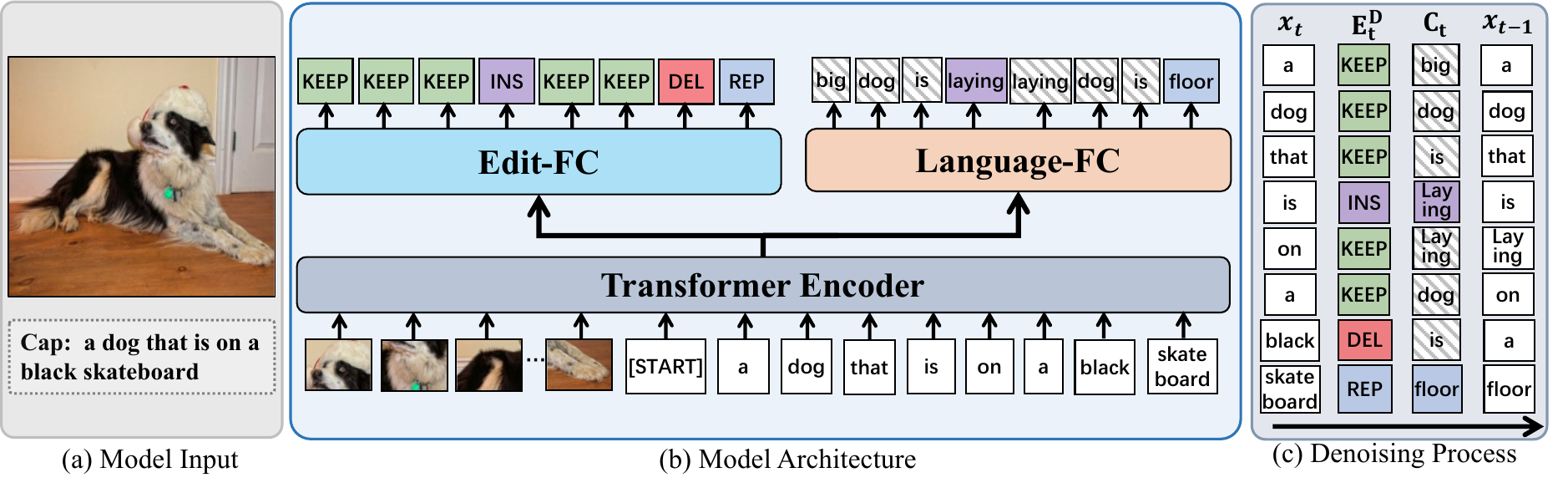}
  \vspace{-1em}
  \caption{The edit-based denoising step and architecture of DECap. DECap will predict a sequence of edit operations and content words to transform the caption. Specifically, contents words are used only when the predicted corresponding edit operation is \texttt{INSERT} or \texttt{REPLACE}, while the rest of the predicted words are abandoned, \ie, the shaded words.}
  \label{fig:model3}
\end{figure*}

\noindent\textbf{Edit-based Denoising Process.}
The edit-based denoising process aims to iteratively edit $\bm{x}_T$ to $\bm{x}_0$ by predicting appropriate edit operations. Specifically, given the image $I$ and the caption $\bm{x}_{t}=\{w^{1}_t, w^{2}_t, ..., w^{l}_t\}$, we model this edit-based denoising process with the explicit prediction of both edit operations and content words transforming $\bm{x}_t$ to $\bm{x}_{t-1}$:
\begin{equation*}
\small
p_{\theta}(\bm{x}_{t-1}|\bm{x}_t, t, I) = p(\bm{x}_{t-1}|\bm{x}_t, E^D_t, C_t) \cdot p( E^D_t, C_t|\bm{x}_t, t, I),
\end{equation*}
where $p_{\theta}$ parameterized the model to predict a sequence of denoising edit operations $E^{D}_t = \{e^1_t, e^2_t, ..., e^l_t\}$, together with a sequence of content words $C_t = \{c^1_t, c^2_t, ..., c^l_t\}$ which all have the same length with $\bm{x}_t$. As the example shown in Fig.~\ref{fig:model3}, the denoising step transforms the caption $\bm{x}_t$ to $\bm{x}_{t-1}$ based on the edit operations and predicted words, \ie, for each word $w^{i}_t$, we keep the original word if it is predicted operation $e^i_t$ is \texttt{KEEP}, remove the word if it is predicted operation $e^i_t$ is \texttt{DELETE}, copy the original word and add a new word $c^i_t$ after it if predicted operation $e^i_t$ is \texttt{INSERT}, and replace it with a new word $c^i_t$ if predicted operation $e^i_t$ is \texttt{REPLACE}. We then feed the output of this step into the model and perform the next denoising step. Following this, we can generate the caption by performing a series of denoising steps from an all random word sequence:
\begin{equation}
\small
P_\theta(\bm{x}_{0})=\textstyle{\prod}_{t=1}^Tp_\theta(\bm{x}_{t-1} | \bm{x}_t, t, I).
\end{equation}

\subsection{Transformer-based Model Architecture}
\label{method:model}

The DECap is built based on the standard Transformer~\cite{vaswani2017attention} architecture, which has strong representation encoding abilities. To facilitate the denoising process, we further construct DECap with a parallelized system for the efficient generation of both edit operations and content words.

\noindent\textbf{Feature Extraction.}  Given an image $I$ and caption $x_t$, we construct the input for the model as a sequence of visual tokens and word tokens. Specifically, we encode the image $I$ into visual tokens through pre-trained visual backbones such as CLIP. The word tokens are represented by the sum of word embedding, position encoding, and segment encoding. Meanwhile, following previous works~\cite{ho2020denoising,austin2021structured,li2022diffusion}, we encode the time step $t$ as a sinusoidal embedding the same way as the position encoding, adding it to the word tokens.

\noindent\textbf{Model Architecture.} As shown in Fig.~\ref{fig:model3}, given the visual-word token sequence with a connecting token, \eg, \texttt{[START]}, we first utilize the Transformer encoder blocks with self-attention and co-attention layers to learn the multi-modal representations of each token. We then use two simple yet effective FC layers to predict the edit operation and content word for each word token. Specifically, by feeding the hidden states of word tokens as input, 1) the Edit-FC generates the edit operation sequence $E^D_t$ by making a four-category classification for each word, \ie, $e_t \in \{\texttt{REPLACE,DELETE,INSERT,KEEP}\}$. 2) In parallel, the Language-FC maps each hidden state to a distribution over the vocabulary to predict specific words to generate the content word sequence $C_t$. Following the denoising step in Sec.~\ref{method:discrete_diffusion}, we then transform the caption $x_t$ to $x_{t-1}$ based on the edit operations and content words for next step.

\subsection{Training Objectives and Inference}
\label{method:train_infer}

\noindent\textbf{Training.} Following previous discrete diffusion works~\cite{austin2021structured,zhu2022exploring}, we train the model to directly predict the original ground-truth caption $\bm{x}_0$ for caption $x_t$:
\begin{equation}
\small
    \mathcal{L} = \mathcal{L}_{Edit} + \mathcal{L}_{Languge} = -\log p_{\theta}(E^G_t|\bm{x}_t,t,I) +  -\log p_{\theta}(C^G_t|\bm{x}_t,t,I),
\end{equation}
where $E^G_t$ and $C^G_t$ are ground truth edit operations and content words constructed based on the $\bm{x}_0$ and $\bm{x}_t$. $\mathcal{L}_{Edit}$ and $\mathcal{L}_{Languge}$ are cross-entropy loss over the distribution of predicted edit operations and content words, and $\mathcal{L}_{Languge}$ is only trained to predict content words for the input words assigned with \texttt{INSERT} and \texttt{REPLACE} operations.

\noindent\textbf{Inference.} Given image and caption $\bm{x}_t$ ($t\in (t,T]$), the model predicts $\bm{x}_{t-1}$, $\bm{x}_{t-2}$ iteratively for $t$ denoising steps, and produces the final result of $\bm{x}_0$.

\section{Experiments}

\subsection{Experimental Setup}

\noindent\textbf{Datasets.} We evaluated DECap on both popular caption editing (\ie, \textbf{COCO-EE}~\cite{wang2022explicit}, \textbf{Flickr30K-EE}\footnote{Due to the limited space, more details are left in the Appendix. \label{ft:appendix}}) and caption generation benchmarks (\ie, \textbf{COCO}~\cite{lin2014microsoft}). Specifically, COCO-EE contains 97,567 training samples, 5,628 validation samples, and 5,366 test samples, where each editing instance consists of one image and one corresponding Ref-GT caption pair. COCO contains 123,287 images with 5 human-annotated captions for each. In this paper, we utilized the Karpathy splits~\cite{karpathy2015deep}, which contain 113,287 training images, 5,000 validation images, and 5,000 test images.

\noindent\textbf{Evaluation Metrics.} We utilized all the prevalent accuracy-based metrics following prior works, which include BLEU-N~\cite{papineni2002bleu}, METEOR~\cite{banerjee2005meteor}, ROUGE-L~\cite{lin2004rouge}, CIDEr-D~\cite{vedantam2015cider}, and SPICE~\cite{anderson2016spice}. Meanwhile, we also computed CLIP-Score~\cite{hessel2021clipscore} to evaluate the caption-image similarity, and the inference time to evaluate the model efficiency.

\begin{table*}[t]
  \renewcommand\arraystretch{1.05}
  \begin{center}
    \scalebox{0.95}{
    \begin{tabular}{l|c|c|cccccccc|c}
    \hline
    \multirow{2}{*}{Model} & {Unpaired} &\multirow{2}{*}{Step} &\multicolumn{8}{c|}{Quality Evaluation} & {Inference}
    \\
    & {Data} & & {B-1} & {B-2} & {B-3} & {B-4} & {R} & {C} & {S} & CLIP-Score &{Time(ms)}\\
    \hline
    Ref-Caps & {---} & {---}& 50.0  & 37.1  & 27.7  & 19.5  & 48.2  & 129.9  & 18.9  & 0.6997 & {---} \\
    \hline
    {TIger~\cite{wang2022explicit}} & \textcolor{red}{\xmark} & 4 & {50.3} & {38.5} & {29.4} & {22.3} & {53.1} & {176.7} & {31.4} & 0.7269 & 614.23\\
    {TIger-N~\cite{wang2022explicit}} & \textcolor{mygreen}{\cmark} & 4 & {51.8} & {38.6} & {28.9} & {20.7} & {49.6} & {145.0} & {21.8} & 0.7097 & 611.72\\
    \textbf{DECap} & \textcolor{mygreen}{\cmark} & 4 & \cellcolor{mygray-bg}{55.5} & \cellcolor{mygray-bg}{41.7} & \cellcolor{mygray-bg}{31.5} & \cellcolor{mygray-bg}{23.3} & \cellcolor{mygray-bg}{52.7} & \cellcolor{mygray-bg}{173.7} & \cellcolor{mygray-bg}{29.8} & \cellcolor{mygray-bg}{0.7439} & \cellcolor{mygray-bg}{277.30}\\
    \textbf{DECap} & \textcolor{mygreen}{\cmark} & 5 & \cellcolor{mygray-bg}{56.0} & \textbf{\cellcolor{mygray-bg}{42.0}} & \textbf{\cellcolor{mygray-bg}{31.6}} & \textbf{\cellcolor{mygray-bg}{23.5}} & \cellcolor{mygray-bg}{53.0} & \cellcolor{mygray-bg}{176.2} & \cellcolor{mygray-bg}{31.4} & \cellcolor{mygray-bg}{0.7498} & \cellcolor{mygray-bg}{335.45} \\
    \textbf{DECap} & \textcolor{mygreen}{\cmark} & 6 & \cellcolor{mygray-bg}{\textbf{56.1}} & \cellcolor{mygray-bg}{41.9} & \cellcolor{mygray-bg}{31.4} & \cellcolor{mygray-bg}{23.4} & \cellcolor{mygray-bg}{\textbf{53.1}} & \cellcolor{mygray-bg}{\textbf{177.0}} & \cellcolor{mygray-bg}{\textbf{32.2}} & \cellcolor{mygray-bg}{\textbf{0.7522}} & \cellcolor{mygray-bg}{409.99} \\
    \hline
    \end{tabular}%
  }
  \end{center}
  \vspace{-1em}
  \caption{The \emph{in-domain} evaluation on the COCO-EE test set. All ECE models were trained on the COCO-EE training set. ``Ref-Caps" denotes the initial quality of given reference captions. ``TIger-N'' denotes the TIger trained with noised unpaired data.}
  \label{tab:in_cocoee}%
\end{table*}%

\subsection{Generalization Ability in ECE}
\label{sec:generalization_ability}

In this subsection, we evaluated the generalization ability of our model with both in-domain and out-of-domain evaluation on the COCO-EE. Specifically, we try to answer four research questions: 1) \textbf{Q1:} Does DECap perform well on the existing in-domain benchmark? 2) \textbf{Q2:} Does DECap perform well on reference captions with different noisy levels (\ie, Levenshtein ratios\footref{ft:ratio})? \textbf{Q3:} Does DECap can further boost the performance of ``good'' reference captions from other captioning models? 3) \textbf{Q4:} Does DECap perform well on pure random reference captions?  It is worth noting that DECap only used the unpaired data (\ie, image and GT-Cap without Ref-Cap) while the state-of-the-art TIger~\cite{wang2022explicit} was trained with the complete editing instance. For a more fair comparison, we also trained TIger with the synthesized noised unpaired data (denoted as TIger-N).


\textbf{In-Domain Evaluation: COCO-EE (Q1)}

\noindent\textbf{Settings.} Since the COCO-EE dataset was carefully designed to emphasize the refinement of content details, its training set and test set have similar distribution (\ie, \texttt{Ratio}\footref{ft:ratio} around 0.5 for most instances), thus we directly compared the performance of each model on the COCO-EE test set as the in-domain evaluation. We evaluated edited captions against their single GT-Cap.

\noindent\textbf{Results.} The in-domain evaluation results are reported in Table~\ref{tab:in_cocoee}. From the table, we can observe: 1) For the quality evaluation, TIger achieves its best performance using four editing steps, while our DECap achieves competitive results with the same step (\eg, better BLEU scores but slightly lower CIDEr-D score). With more editing steps, DECap can further improve the quality of captions, outperforming TIger on all metrics. \emph{It is worth noting that TIger was even trained on the in-domain Ref-GT caption pairs}. 2) TIger-N achieves limited quality improvement on the in-domain samples. 3) For the efficiency evaluation, DECap achieves significantly faster inference speed than TIger even with more editing steps. This is because DeCap predicts edit operations and content words simultaneously but TIger needs to conduct editing by three sequential modules.

\begin{figure*}[!t]
  \centering
  \vspace{-1em}
  \includegraphics[width=1\linewidth]{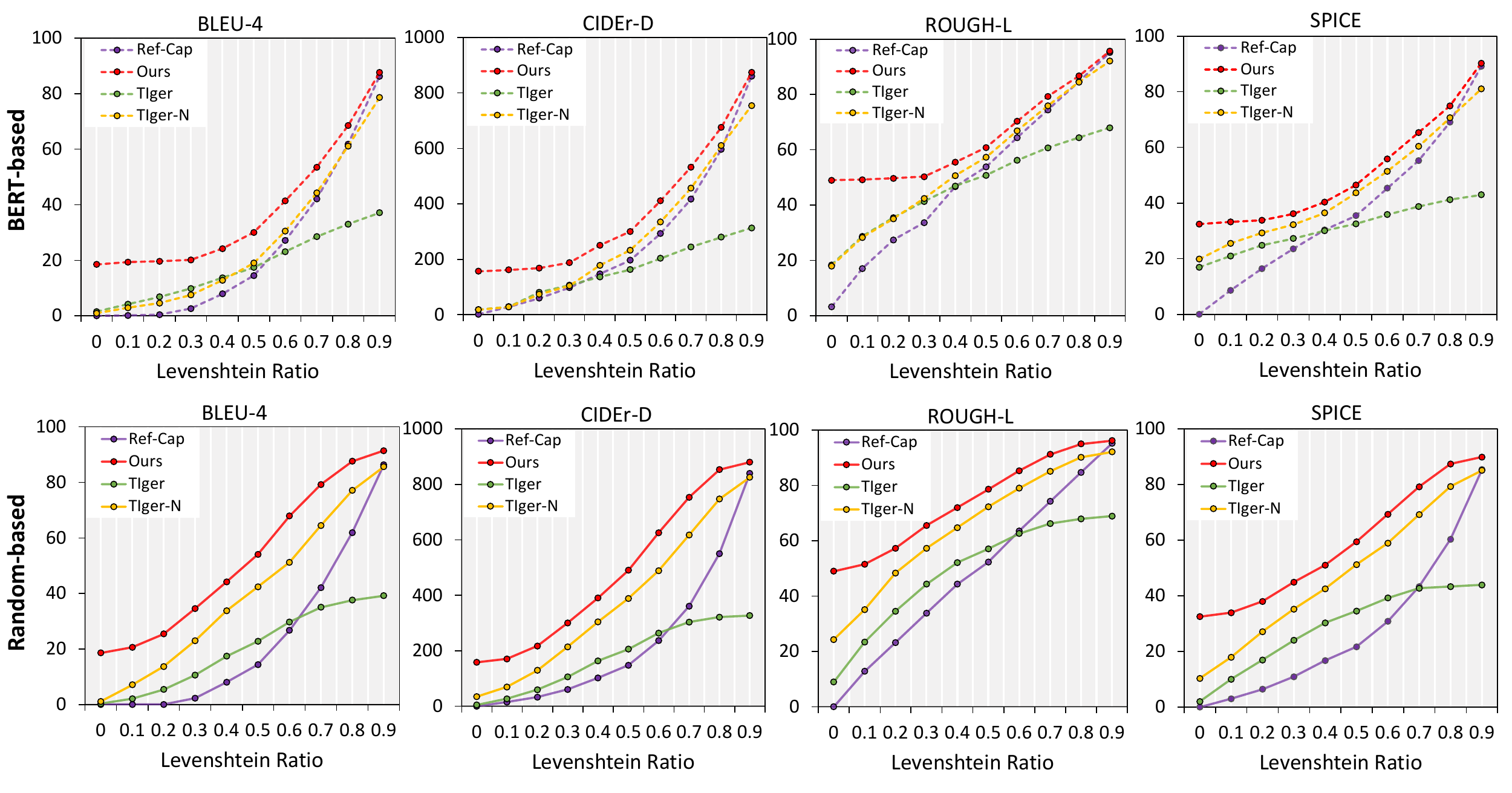}
  \vspace{-1.5em}
  \caption{Performance on two kinds of out-of-domain GT-based reference captions constructed from COCO-EE test set. All models were trained on the COCO-EE training set. ``Ref-Caps" denotes the initial quality of given reference captions, and ``TIger-N'' denotes the TIger trained with unpaired data.}
  \label{fig:out_domain1}
  \vspace{-1.5em}
\end{figure*}

\begin{table}[t]
  \begin{center}
  \setlength{\tabcolsep}{1.8mm}
    \scalebox{0.99}{
    \begin{tabular}{l|cccccccc}
    \hline
    \multirow{2}{*}{Model} &\multicolumn{8}{c}{Quality Evaluation}\\
    & {B-1} & {B-2} & {B-3} & {B-4} & {R} & {C} & {S} & {CLIP-Score}\\
    \hline
    {Up-Down~\cite{anderson2018bottom}} & 74.9 & 58.6 & 45.2 & {35.1} & {55.9} & {109.9} & {20.0} & 0.7359\\
    {+ TIger~\cite{wang2022explicit} \textcolor{red}{$\downarrow$}} & 70.3 &  54.9 & 41.2 &  {30.9} & {54.3} & {95.7} & {17.1} &  0.7285\\
    {+ TIger-N~\cite{wang2022explicit} \textcolor{red}{$\downarrow$}} & 74.8 & 58.5 & 45.1 &  {34.8} & {55.8} & {109.5} & {19.9} &  0.7356\\
    {+ \textbf{DECap} \textcolor[RGB]{66,185,131}{$\uparrow$}} & \cellcolor{mygray-bg}{\textbf{75.1}} & \cellcolor{mygray-bg}{\textbf{58.9}} & \cellcolor{mygray-bg}{\textbf{45.5}}& \cellcolor{mygray-bg}{\textbf{35.2}} & \cellcolor{mygray-bg}{\textbf{56.3}} & \cellcolor{mygray-bg}{\textbf{112.3}} & \cellcolor{mygray-bg}{\textbf{20.2}} &  \cellcolor{mygray-bg}{\textbf{0.7432}} \\
    \hline
    {Transformer~\cite{sharma2018conceptual}} &  75.2 & 58.9 & 45.6 & {35.5} & {56.0} & {112.8} & {20.6} & 0.7452\\
    {+ TIger~\cite{wang2022explicit} \textcolor{red}{$\downarrow$}} & 70.0 & 55.0 & 41.5 & {31.1} & {54.4} & {97.1} & {17.3} &  0.7313\\
    {+ TIger-N~\cite{wang2022explicit} \textcolor{red}{$\downarrow$}} &  75.1 & 58.8 & 45.4 & {35.2} & {55.9} & {112.4} & {20.4} &  0.7443\\
    {+ \textbf{DECap} \textcolor[RGB]{66,185,131}{$\uparrow$}} & \cellcolor{mygray-bg}{\textbf{75.5}} & \cellcolor{mygray-bg}{\textbf{59.2}}& \cellcolor{mygray-bg}{\textbf{45.8}} & \cellcolor{mygray-bg}{\textbf{35.6}} & \cellcolor{mygray-bg}{\textbf{56.2}} & \cellcolor{mygray-bg}{\textbf{114.1}} & \cellcolor{mygray-bg}{\textbf{20.7}} & \cellcolor{mygray-bg}{\textbf{0.7472}}\\
    \hline
    {BLIP~\cite{li2022blip}} & 79.7 & 64.9 & 51.4 & {40.4} & {60.6} & {136.7} & {24.3} & 0.7734\\
    {+ TIger~\cite{wang2022explicit} \textcolor{red}{$\downarrow$}} & 73.5 & 58.7 & 44.7 & {33.7} & {56.3} & {104.0} & {18.2} &   0.7377\\
    {+ TIger-N~\cite{wang2022explicit} \textcolor{red}{$\downarrow$}} & 79.5 & 64.5 & 51.0 &  {39.9} & {60.3} & {133.0} & {23.7} &   0.7690\\
    {+ \textbf{DECap} \textcolor[RGB]{66,185,131}{$\uparrow$}} & \cellcolor{mygray-bg}{\textbf{79.9}} & \cellcolor{mygray-bg}{\textbf{65.1}} & \cellcolor{mygray-bg}{\textbf{51.7}} &  \cellcolor{mygray-bg}{\textbf{40.6}} & \cellcolor{mygray-bg}{\textbf{60.7}} & \cellcolor{mygray-bg}{\textbf{138.1}} & \cellcolor{mygray-bg}{\textbf{24.4}} & \cellcolor{mygray-bg}{\textbf{0.7738}}\\
    \hline
    \end{tabular}%
  }
  \end{center}
  \vspace{-1em}
  \caption{Performance of ECE models editing model-generated captions on COCO test set. All ECE models were trained on the COCO-EE training set. ``TIger-N'' denotes the TIger trained with unpaired data.}
  \vspace{-2em}
  \label{tab:out3_cocoee}%
\end{table}%

\begin{table*}[t]
  \renewcommand\arraystretch{1.06}
  \begin{center}
    \scalebox{0.95}{
    \begin{tabular}{l|c|c|cccccccc|c}
    \hline
    \multirow{2}{*}{Model} & {Unpaired} &\multirow{2}{*}{Step} &\multicolumn{8}{c|}{Quality Evaluation} & {Inference}\\
    & {Data} & & {B-1} & {B-2} & {B-3} & {B-4} & {R} & {C} & {S} & {CLIP-Score} &{Time(ms)}\\
    \hline
    {TIger~\cite{wang2022explicit}} & \textcolor{red}{\xmark} & 10 & {14.7} & {4.6} & {1.9} & {0.9} & {13.5} & {3.0} & {1.2} & 0.5158 & 1413.16\\
    {TIger-N~\cite{wang2022explicit}} & \textcolor{mygreen}{\cmark} & 10& 7.2 & {5.9} & {4.5} & {3.5} & {29.1} & {23.5} & {4.8} & 0.6116 & 1417.09\\
    \textbf{DECap} & \textcolor{mygreen}{\cmark} & 10 & \cellcolor{mygray-bg}{\textbf{74.7}} & \cellcolor{mygray-bg}{\textbf{57.4}} & \cellcolor{mygray-bg}{\textbf{42.1}} & \cellcolor{mygray-bg}{\textbf{30.0}} & \cellcolor{mygray-bg}{\textbf{55.3}} & \cellcolor{mygray-bg}{\textbf{102.5}} & \cellcolor{mygray-bg}{\textbf{19.6}} & \cellcolor{mygray-bg}{\textbf{0.7501}} & \cellcolor{mygray-bg}{684.32}  \\
    \hline
    \end{tabular}%
  }
  \end{center}
  \vspace{-1em}
  \caption{Performance of ECE models on pure random (ten random words) reference captions constructed based on the COCO test set. All models were trained on the COCO-EE training set. ``TIger-N'' denotes TIger trained with noised unpaired data.}
  \label{tab:out2_cocoee}%
\end{table*}%

\textbf{OOD: GT-Based Reference Caption (Q2)}

\noindent\textbf{Setting.}  The GT-based reference captions were constructed based on the GT-Caps in the COCO-EE test set. We systematically replaced words in GT-Caps with other words, resulting in the creation of various out-of-domain Ref-Caps. They varied in terms of their Levenshtein ratios, ranging from 0.9 (\ie, with only a few incorrect words) to 0.0 (\ie, where all words were wrong). Specifically, we constructed two kinds of GT-based reference captions: 1) \textbf{BERT-based}. We first replaced the GT words with the special \texttt{[MASK]} tokens and then utilized the pretrained BERT~\cite{devlin2018bert} model to predict other words different from GT words. 2) \textbf{Random-based}. We directly replaced GT words with other random words. We evaluated edited captions against their single GT-Cap.

\noindent\textbf{Results.} As shown in Fig.~\ref{fig:out_domain1}, For models trained with unpaired data, our model successfully improves the quality of all kinds of the GT-based Ref captions (\ie, BERT- and Random-based) and surpasses TIger-N. In contrast, TIger struggles when editing Ref captions with either ``minor'' or ``severe'' errors, and even degrading the captions' quality (\eg, Ref-Caps with ratio larger than 0.6\footref{ft:appendix}) by inadvertently removing accurate words or failing to introduce accurate details.

\textbf{OOD: Model-Generated Reference Caption (Q3)}

\noindent\textbf{Setting.} We explored the models' generalization ability to further improve the quality of captions generated by captioning models. We utilized captions generated by effective captioning models~\cite{anderson2018bottom, sharma2018conceptual, li2022blip} on the COCO test set as reference captions. We evaluated the edited captions against their corresponding five GT-Caps. Results are in Table~\ref{tab:out3_cocoee}.

\noindent\textbf{Results.} From Table~\ref{tab:out3_cocoee}, we can observe: 1) Our model successfully improves the quality of captions generated by captioning models. 2) Both TIger and TIger-N fail to do so and even degrading the caption's quality. 3) Notebly, while existing captioning models fall short of achieving comparable performance with the powerful vision-language pretrained models (\eg, BLIP~\cite{li2022blip}), our DECap trained solely on COCO-EE, demonstrates its unique editing ability to further enhance the quality of captions generated by BLIP.

\textbf{OOD: Pure Random Reference Caption (Q4)}

\noindent\textbf{Setting.} To further evaluate the models' generalization ability without utilizing any GT captions, we constructed pure random reference captions based on the COCO test set. Specifically, each editing instance consists of a single image and a Ref-Cap with ten random words. We evaluated the edited captions against their corresponding five GT-Caps. All results are reported in Table~\ref{tab:out2_cocoee}.

\noindent\textbf{Results.} From Table~\ref{tab:out2_cocoee}, we can observe: 1) Given the image, all the models achieve their best performance with ten editing steps, DECap successfully edits the sentence with all random words into a coherent caption. In contrast, both TIger and TIger-N face challenges in doing so. 2) For efficiency metrics, DECap achieves significantly faster inference speed compared to TIger and TIger-N.

\subsection{Conventional Caption Generation Ability}
\label{sec:caption_generation}
Surprised by the results of editing model-generated captions and pure random reference captions, we further investigated DECap's capacity for directly generating captions.

\noindent\textbf{Settings.} We compared DECap with SOTA diffusion-based captioning approaches on the COCO dataset, especially the discrete diffusion-based captioning model DDCap~\cite{zhu2022exploring}. We trained DECap on the COCO training set with a vocabulary size of 23,531 together with different diffusion steps (10 and 15). During testing, we constructed input instances consisting of a single image from the COCO test set and a Ref-Cap with ten random words. The edited captions were then evaluated against the corresponding five GT-Caps.

\noindent\textbf{Results.} From Table~\ref{tab:coco}, we can observe: 1) Within ten editing steps, DECap achieves superior performance on key metrics (\eg, CIDEr-D and SPICE) compared with other diffusion-based captioning works which even need more diffusion steps. While it falls slightly behind SCD-Net on BLEU-N, it's important to note that CIDEr and SPICE metrics are specifically designed for captioning evaluation and are better aligned with human judgments (than BLEU-N). 2) DECap achieves a significantly faster inference speed compared with another discrete diffusion-based model DDCap (675.80 vs. 3282.58ms). 3) Additionally, DECap can further boost the performance with more editing steps (\eg, 121.2 on CIDEr-D with 15 steps) and keep the inference speed at a reasonable level (\ie, 933.10ms). These results suggest the remarkable potential of DECap in improving the quality of caption generation in a more explainable and efficient edit-based manner.

\begin{table}[t]
\vspace{0.5em}
  \renewcommand\arraystretch{1.06}
  \begin{center}
    \setlength{\tabcolsep}{1.8mm}
    \scalebox{0.99}{
    \begin{tabular}{l|c|ccccccccccc}
    \hline
    Model & Step & {B-1} &{B-2} & {B-3}&  {B-4} & {M} & {R} & {C} & {S} \\
    \hline
    \multicolumn{10}{l}{\emph{Continuous Diffusion}}\\
    \hline
 {Bit Diffusion~\cite{chen2022analog}} & 20 & --- & --- & --- & 34.7 & {---} &{58.0}  & 115.0  & {---} \\
   {SCD-Net~\cite{luo2023semantic}} & 50  & \textcolor{blue}{\textbf{79.0}} & \textcolor{blue}{\textbf{63.4}} & \textcolor{blue}{\textbf{49.1}} & \textcolor{blue}{\textbf{37.3}} & {28.1} &{58.0}  & 118.0  & {21.6} \\
    \hline
    \multicolumn{10}{l}{\emph{Discrete Diffusion}}\\
    \hline
 {DDCap~\cite{zhu2022exploring}} & 20  & --- & --- & --- & 35.0 & {28.2} &{57.4}  & 117.8  & {21.7}  \\
   \textbf{DECap}& 10  & \cellcolor{mygray-bg}{{78.0}} & \cellcolor{mygray-bg}{{61.4}} & \cellcolor{mygray-bg}{{46.4}} & \cellcolor{mygray-bg}{34.5} & \cellcolor{mygray-bg}{\textcolor{red}{28.6}} & \cellcolor{mygray-bg}{\textcolor{red}{58.0}}  & \cellcolor{mygray-bg}{\textcolor{red}{119.0}}  & \cellcolor{mygray-bg}{\textcolor{red}{21.9}} \\
   \textbf{DECap}& 15  & \cellcolor{mygray-bg}{\textcolor{red}{{78.5}}} & \cellcolor{mygray-bg}{\textcolor{red}{{62.2}}} & \cellcolor{mygray-bg}{\textcolor{red}{{47.4}}} & \cellcolor{mygray-bg}{\textcolor{red}{35.3}} &  \cellcolor{mygray-bg}{\textcolor{blue}{\textbf{29.0}}} &  \cellcolor{mygray-bg}{\textcolor{blue}{\textbf{58.4}}} & \cellcolor{mygray-bg}{\textcolor{blue}{\textbf{121.2}}}  & \cellcolor{mygray-bg}{\textcolor{blue}{\textbf{22.7}}} \\
    \hline
    \end{tabular}%
  }
  \end{center}
  \vspace{-0.5em}
  \caption{Comparison between our DECap and state-of-the-art diffusion-based captioning models on the COCO test set. All models were trained on the COCO training set. The \textcolor{blue}{\textbf{best}} and \textcolor{red}{second best} results are denoted with corresponding formats.}
  \label{tab:coco}%
\end{table}%

\subsection{Potential Ability: Controllable Captioning}

Building on the remarkable generalization ability exhibited by DECap in both caption editing and generation, we further conducted a preliminary exploration of its potential for controllability. Compared to existing CIC methods, which offer only coarse control over contents and structures, we can achieve precise and explicit control over caption generation through predefined control words.

\noindent\textbf{Settings.} We constructed input instances consisting of a single image from the COCO test set and a sentence with ten random words. We replaced several random words with specific control words (\eg, objects and attributes) at predefined positions based on the visual information of images.

\begin{figure*}[t]
    \begin{minipage}{0.53\linewidth}
    \includegraphics[width=\columnwidth]{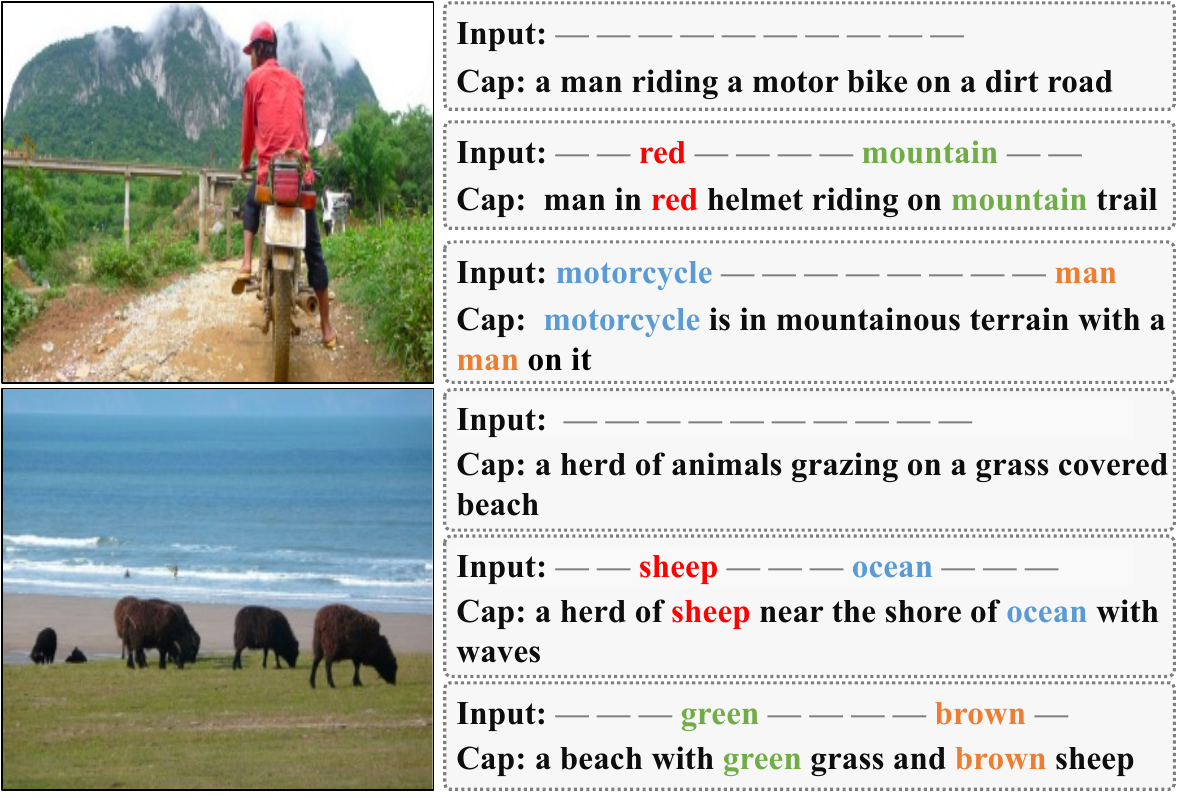}
  \vspace{-1.5em}
  \caption{Controllability of DECap. The grey lines represent random words from the vocabulary, and other colored words represent the manually placed control words.}
  \label{fig:contorl}
  \vfill
    \end{minipage}
\hfill
    \begin{minipage}{0.45\linewidth}
  \begin{minipage}[t]{\textwidth}
\setlength{\tabcolsep}{1.8mm}
    \scalebox{0.85}{
       \begin{tabular}{c|ccccc}
    \hline
    {RW} & {B-4} & {M} & {R} & {C} & {S}
    \\
    \hline
    8 & {32.4} &  26.3 & {56.5} & {109.7} & {20.1}\\
    9 & \textbf{{35.5}} & 27.8 & {58.0} & {118.1} & {21.5} \\
    10 & 34.5 & {28.6} & \textbf{{58.0}}  & \textbf{{119.0}}  & {21.9}\\
    11 &  {32.9} & 28.9 & 57.1 & 115.7 & 22.4\\
    12 & {31.2} & \textbf{29.0} & 56.1 & 109.3 & \textbf{22.7}\\
    \hline
    \end{tabular}%
    }
    \vspace{-1.0em}
    \captionof{table}{Performance of DECap on the COCO test set with different numbers of input random words (RW).}
  \label{tab:coco_num}%
  \end{minipage}
  
\vspace{1em}
  
\begin{minipage}[t]{\textwidth}
\setlength{\tabcolsep}{0.9mm}
    \scalebox{0.85}{
       \begin{tabular}{c|ccccc}
    \hline
    {Distribution} & {B-4} & {M} & {R} & {C} & {S}
    \\
    \hline
    $\alpha = \beta = \gamma$ & {34.0} & 28.6 & 57.8 & 117.4 & 21.8\\
    $\alpha > \beta = \gamma$ & \textbf{34.5}& \textbf{28.6} & \textbf{{58.0}}  & \textbf{{119.0}}  & \textbf{21.9}\\
    $\beta = \gamma = 0$ & {{33.8}} & 25.8 & {57.8} & {116.6} & {21.7} \\
    \hline
    \end{tabular}%
    }
    \vspace{-1.0em}
    \captionof{table}{Performance of DECap on the COCO test set when with different distributions of noising edit types.}
  \label{tab:coco_edit_type}%
  \end{minipage}
  
    \end{minipage}
\end{figure*}

\noindent\textbf{Results.} As shown in Fig.~\ref{fig:contorl}, DECap is capable of editing sentences based on input control words, \ie, all generated captions follow the order of the given control words with guaranteed fluency. Meanwhile, DECap shows its reasoning ability to generate relevant semantic content based on the control words: 1) Given the attributes (\eg, color), DECap can generate specific contents with these attributes (\eg, ``\textcolor{red}{\texttt{red}}" $\rightarrow$ ``\texttt{helmet}", ``\textcolor[RGB]{66,185,131}{\texttt{green}}" $\rightarrow$ ``\texttt{grass}" and ``\textcolor{brown}{\texttt{brown}}" $\rightarrow$ ``\texttt{sheep}"). 2) Given objects, DECap can generate further descriptions or related objects (\eg, ``\texttt{mountain}" $\rightarrow$ ``\texttt{trail}" and ``\texttt{ocean}" $\rightarrow$ ``\texttt{wave}"). These results indicate the potential of DECap to enhance controllability and diversity, achieving a more direct and word-level control beyond existing CIC methods.

\subsection{Ablation Study}
\label{sec:caption_ablations}

\noindent\textbf{Number of Random Words.} In this section, we run a set of ablations about the influence of different numbers of random words on caption generation. We utilized the DECap trained with diffusion step $T=10$ from Sec.~\ref{sec:caption_generation}, and constructed input instances consisting of an image from the COCO test set and a Ref-Cap with $n$ random words, where $n \in \{ 8,9,10,11,12\}$. As shown in Table~\ref{tab:coco_num}, DECap's performance consistently improves as the number of random words increases from 8 to 10 and starts to decline beyond 10\footref{ft:appendix}. DECap achieves its highest CIDEr-D score when editing sentences with 10 random words as the average length of GT captions in COCO is around 10. We thus selected 10 words as a balanced choice for caption generation.

\noindent\textbf{Distribution of Edit Types.} As discussed in Sec.\ref{method:discrete_diffusion}, the distribution over edit types plays a crucial role in balancing different noising operations and training diverse denoising abilities. Therefore, we examined the impact of varying distribution settings for the edit types within the edit-based noising process. Specifically, the probabilities for the noising edit operations \texttt{REPLACE}, \texttt{DELETE}, and \texttt{INSERT} are denoted as $\alpha$, $\beta$, and $\gamma$, respectively. We perform ablations by imposing global control over these probabilities\footref{ft:appendix}. We trained the DECap on the COCO training set with different distributions of edit types with the same diffusion step $T=10$.  During testing, we constructed input instances consisting of a single image from the COCO test set and a Ref-Cap with ten random words. From Table~\ref{tab:coco_edit_type}, we can observe: 1) DECap performs better when emphasizing the denoising ability of the replacement operation compared to an even distribution of edit types. 2) Training DECap exclusively for the replacement operation, neglecting deletion and insertion abilities, leads to a noticeable decline in caption quality. 3) $\alpha \textgreater \beta$$=$$\gamma$ could be a sensible choice for caption generation. Importantly, our method allows for flexible adaptation, enabling us to set different edit type distributions tailored to specific tasks or requirements. 

\begin{figure}[!t]
  \centering
  \includegraphics[width=0.95
\linewidth]{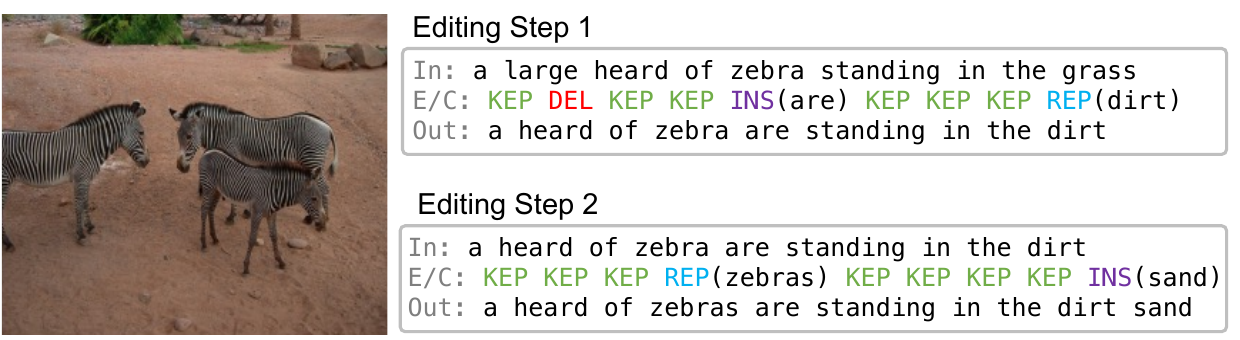}
  \vspace{-1em}
  \caption{Editing process of DECap. ``E'' and ``C'' denote the edit operation and corresponding content word, respectively.}
  \label{fig:edit_vis}
\end{figure}

\noindent\textbf{Visualization.} 
Fig.~\ref{fig:edit_vis} shows a two-step editing example\footref{ft:appendix}.

\section{Conclusion}

In this paper, we pointed out the challenge of limited generalization ability in existing ECE models. And we proposed a novel diffusion-based ECE model, DECap, which reformulates ECE with a discrete diffusion mechanism, incorporating an innovative edit-based noising and denoising process. Extensive results have demonstrated DECap's strong generalization ability and potential as a uniform framework for both caption editing and generation. Moving forward, we are going to: 1) extend DECap into other modalities beyond images, \eg., video; 2) explore advanced techniques for finer controllability of DECap's editing process.

\bibliographystyle{splncs04}
\bibliography{main}

\clearpage  

\begin{center}%
  {\LARGE \textbf{*** Supplementary Manuscript ***} \par}%
\end{center}

\appendix

This supplementary document is organized as follows:
\begin{itemize}
    \item[$\bullet$]  In Sec.~\ref{sec:ratio}, we show more details about the Levenshtein ratio.

    \item[$\bullet$]  In Sec.~\ref{sec:implement}, we show the implementation details.

    \item[$\bullet$]  In Sec.~\ref{sec:visual}, we provide more visualization results.

    \item[$\bullet$]  In Sec.~\ref{sec:lmm}, we provide the results of Multimodal LLMs on ECE.

    \item[$\bullet$]  In Sec.~\ref{sec:dis}, we show the data distribution of the COCO-EE based on the Levenshtein ratio.

    \item[$\bullet$]  In Sec.~\ref{sec:flee}, we show more results about the generalization ability of DECap on the Fickr30K-EE dataset.

    \item[$\bullet$]  In Sec.~\ref{sec:num_of_words}, we provide more detailed ablation study about the number of random words in caption generation.

    \item[$\bullet$]  In Sec.~\ref{sec:edit_type}, we provide more detailed ablation study about the distribution of edit types.


    \item[$\bullet$] Potential Negative Societal Impacts in Sec.~\ref{sec:social}

\end{itemize}

\section{Details for Levenshtein ratio}
\label{sec:ratio}

In this paper, we used the Levenshtein ratio to quantify the similarity between two captions by considering their length and the edit distance needed to transform one into the other. Specifically, for two captions with length $m$ and $n$, the Levenshtein ratio is calculated as:
\begin{equation}
    \texttt{ratio} = \frac{m+n-{ldist}}{m+n} 
\end{equation}
where $ldist$ is the weighted edit distance based on the standard Levenshtein distance~\cite{levenshtein1966binary}. The Levenshtein distance refers to the minimum number of edit operations required to transform one sentence into another, including three Levenshtein operations \texttt{REPLACE}, \texttt{INSERT}, and \texttt{DELETE}. In the case of the weighted version, when calculating $ldist$, both \texttt{INSERT} and \texttt{DELETE} operations are still counted as $+1$, while each \texttt{REPLACE} operation incurs a cost of $+2$:
\begin{equation}
\small
\begin{aligned}
    ldist &= \mathbf{Num}(\texttt{INSERT}) + \mathbf{Num}(\texttt{DELETE}) \\ 
    & + 2*\mathbf{Num}(\texttt{REPLACE})
\end{aligned}
\end{equation}
where $\mathbf{Num(\cdot)}$ represents the number of different edit operations. The range of Levenshtein ratio yields from 0 to 1, where a higher value indicates higher similarity.

\section{Implementation Details.} 
\label{sec:implement}

For image features, we used the ViT features extracted by the ViT-B/16 ~\cite{dosovitskiy2020image} backbone from the pretrained CLIP model~\cite{radford2021learning} with image patch size 16. For the edit-based noising process, we set $\alpha \textgreater \beta$$=$$\gamma$ to emphasize the denoising ability of replacement. For our diffusion model, we used the 12-layer Transformer encoder. We trained our model with Adam optimizer for 50 epochs, and we used a linear decay learning rate schedule with warm up. The initial learning rate was set to 1e-4. Specifically, for Sec.\textcolor{red}{4.2}, we traind DECap with diffusion step $T=10$, and all the three models DECap, TIger and TIger-N use the same vocabulary sized 12,071. The inference time was evaluated as the average run time for each instance on a single A100 GPU with a mini-batch size of 1.

\begin{figure*}[t]
  \centering
  \vspace{0.5em}
  \includegraphics[width=1\linewidth]{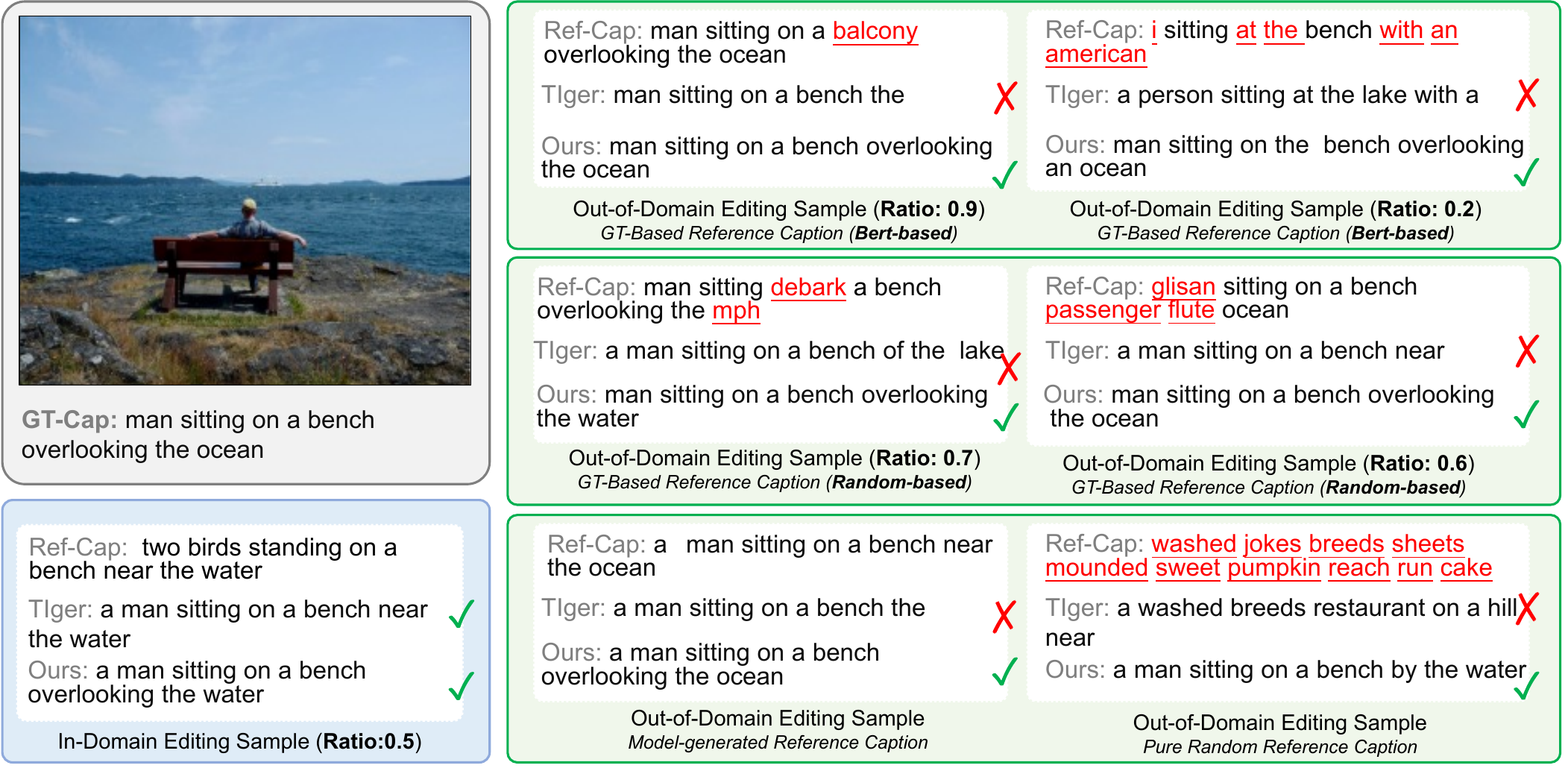}
  \caption{Editing results of state-of-the-art ECE model TIger and our DECap. The \emph{in-domain} Ref-Cap sample is from the COCO-EE test set. The \emph{out-of-domain} Ref-Cap samples include the GT-based reference caption, model-generated reference caption, and pure random reference caption.}
  \label{fig:visual}
\end{figure*}

\section{More Visualization Results}
\label{sec:visual}

\noindent{\textbf{Generalization Ability}.}
As illustrated in Figure~\ref{fig:visual}, existing ECE model TIger~\cite{wang2022explicit} exhibits limited capability in refining \emph{Out-of-Domain} GT-Based reference captions, particularly when the similarity between Ref-Cap and GT-Cap deviates significantly from the balanced value (\ie, \texttt{Ratio} 0.5) seen in training data. In contrast, DECap displays a remarkable generalization ability, successfully editing both \emph{In-Domain} and \emph{Out-of-Domain} samples with diverse editing scenarios.

\begin{figure*}[t]
  \centering
  \includegraphics[width=0.8\linewidth]{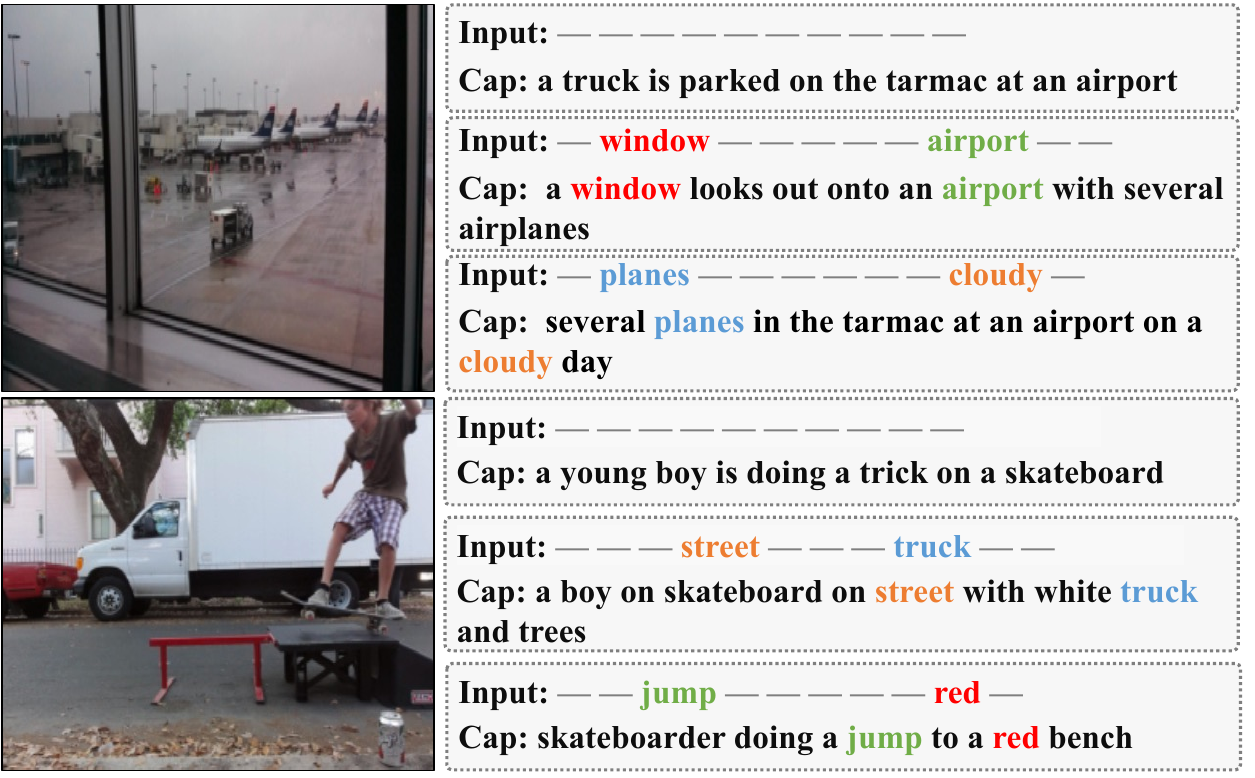}
  \caption{Controllability of DECap. The grey lines represent random words from the vocabulary, other colored words represent the manually placed control words.}
  \vspace{-0.5em}
  \label{fig:visual2}
\end{figure*}

\noindent{\textbf{Potential in Controllability}.} As illustrated in Figure~\ref{fig:visual2}, our model effectively edits sentences based on specific input control words. All the generated captions maintain the order of the provided control words, ensuring both fluency and semantic relevance to these control words.

\begin{figure*}[t]
  \centering
  \vspace{-0.5em}
  \includegraphics[width=1\linewidth]{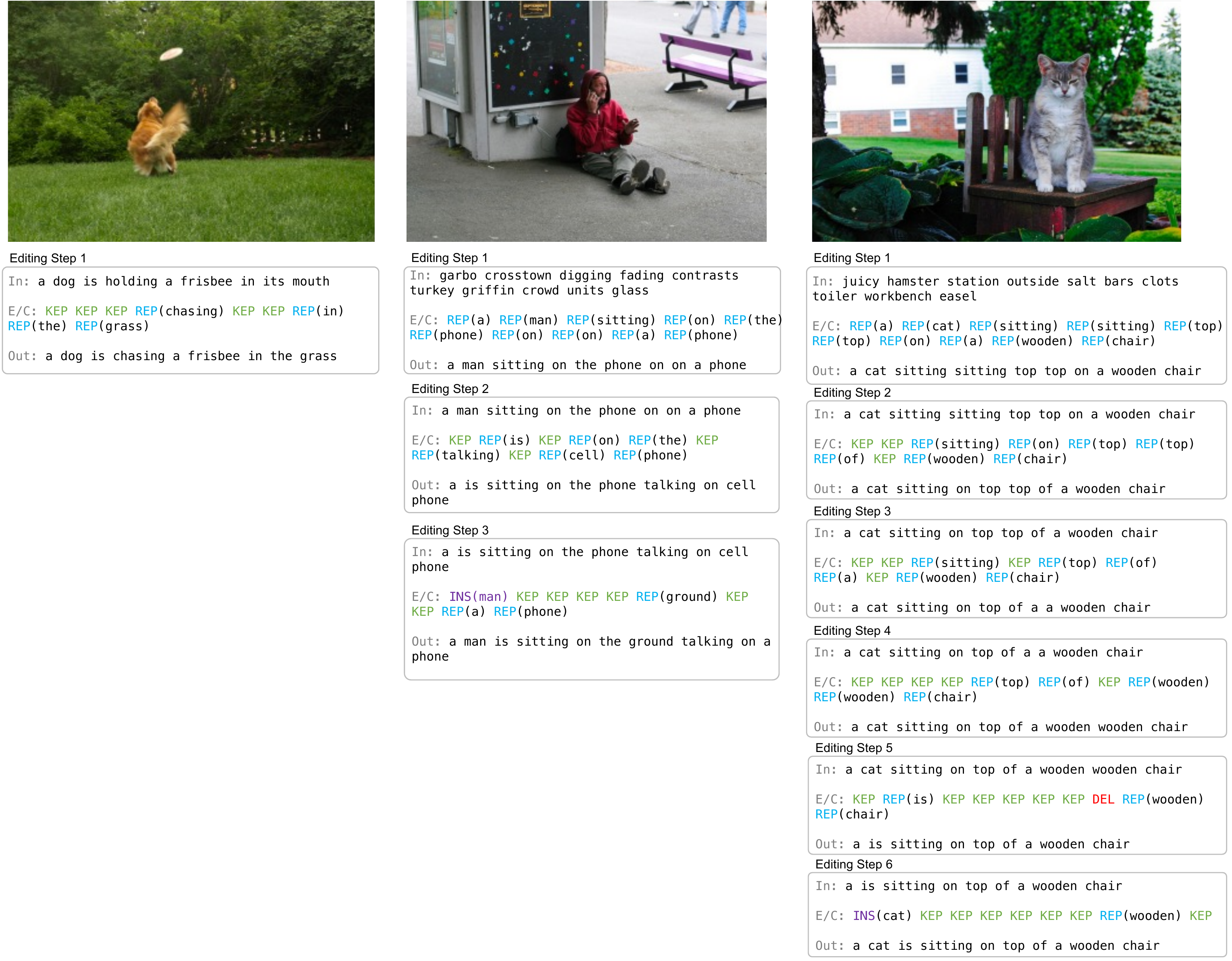}
  \caption{Editing process of DECap. ``E'' and ``C'' denote the edit operation and corresponding content word, respectively.}
  \vspace{-1.5em}
  \label{fig:edit_process}
\end{figure*}

\noindent\textbf{Editing Process.} 
Figure~\ref{fig:edit_process} shows more examples of the editing process of DECap on different editing samples with different editing steps. 

\clearpage

\begin{figure*}[ht]
  \centering
  \vspace{0.5em}
  \includegraphics[width=1\linewidth]{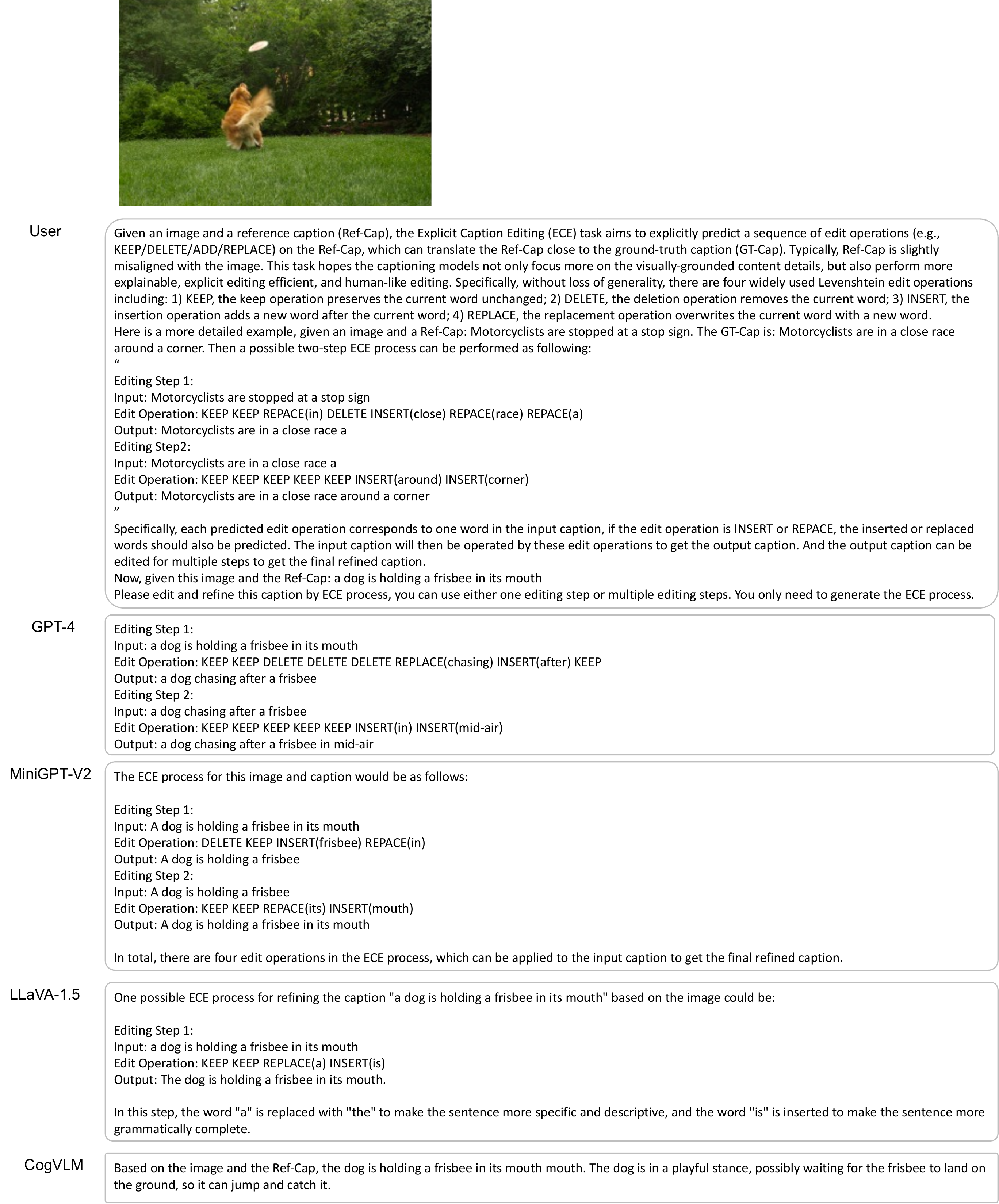}
  \caption{The results of Multimodal LLMs for ECE.}
  \vspace{-1.0em}
  \label{fig:llm}
\end{figure*}

\clearpage

\section{Multimodal LLMs for ECE}
\label{sec:lmm}
In this section, we provide a preliminary exploration of utilizing the multimodal Large Language Models (LLMs) for ECE. 

\noindent{\textbf{Setting}.} We utilized the available opensourced Multimodal LLMs including the GPT-4~\cite{achiam2023gpt}, MiniGPT-v2~\cite{chen2023minigpt}, LLaVA-1.5~\cite{liu2023improved} and CogVLM~\cite{wang2023cogvlm}. The user input prompt consists of the introduction of the ECE task, a definition of different edit operations, an example of a two-step editing process, and finally an editing instance with one image and one Ref-Cap. The multimodal LLMs were then asked to generate the corresponding editing process.

\noindent{\textbf{Results}.} As illustrated in Figure~\ref{fig:llm}, all the multimodal LLMs failed to perform the ECE properly: 1) \textbf{Missalignment of the caption and edit operation}. Normally, the model should generate one edit operation for each word in the input caption correspondingly, thus the sequence of edit operation and input caption should at least have the same length. However, both GPT-4, MiniGPT-v2 and LLaVA-1.5 fail to predict enough edit operations for the input caption. 2) \textbf{Wrong editing transformation}. The edit operation should be operated on each corresponding input word correctly to transform the input caption for the next step. However, both GPT-4, MiniGPT-v2 and LLaVA-1.5 fail to conduct the editing transformation, \eg, MiniGPT-v2 predicts the \texttt{DELETE} operation as the first edit operation, but the corresponding input word ``A" is not deleted, while the phrase ``in its mouth" is deleted without any predicted operations. 3) \textbf{Poor editing ability}. Besides GPT-4, the final output caption of other multimodal LLMs is whether the repetition of the input caption, or just changing the word ``a" into ``the", without any refinement on the misaligned details. The CogVLM even fails to generate the editing process. We argue that these results may related to the ``autoregressive" manner of existing multimodal LLMs, while a ``non-autoregressive" model like DECap is more suitable for this task, which generates edit operation and content word for each input word simultaneously.

\section{Data Distribution of COCO-EE}
\label{sec:dis}
\begin{figure*}
  \centering
  \includegraphics[width=0.8\linewidth]{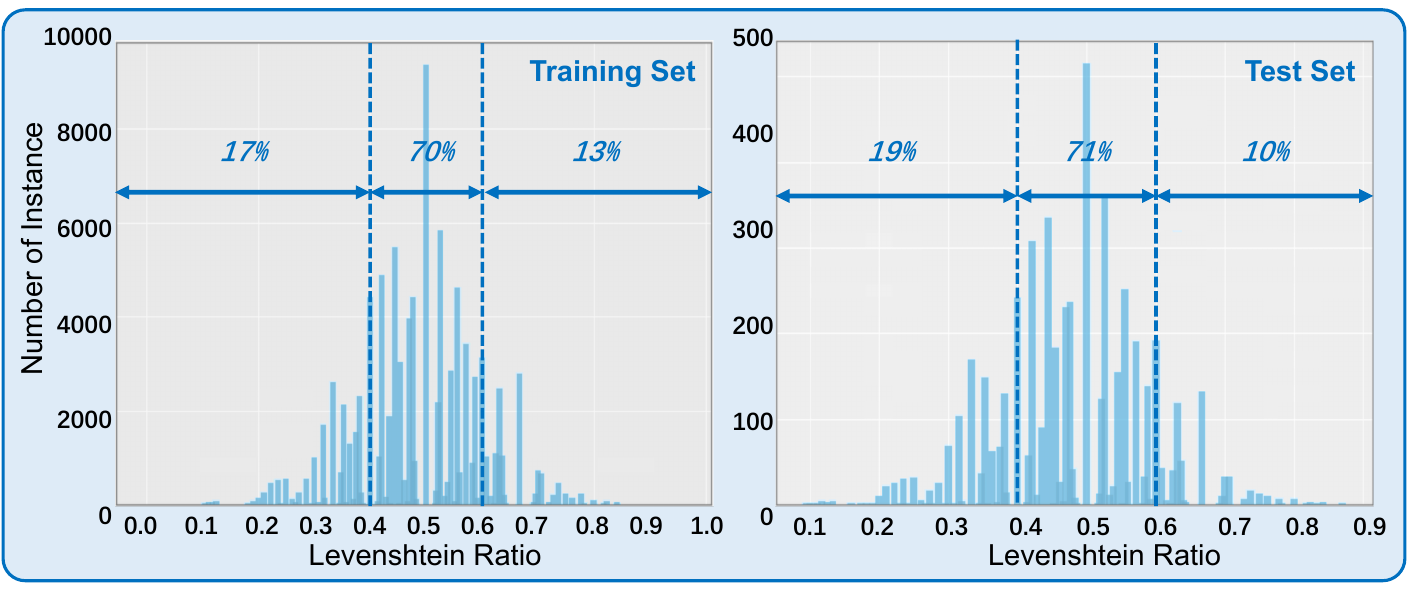}
  \caption{The data distribution of the COCO-EE dataset.}
  \label{fig:dis_coco}
\end{figure*}

As illustrated in Figure~\ref{fig:dis_coco}, we can observe that the training set and test set of COCO-EE have similar distribution: 1) More than 70\% of the editing instances have ratios ranging from 0.4 to 0.6, with the majority of them concentrated around 0.5. 2) There are very few samples with ratios below 0.4 or above 0.6, and almost no samples with ratios around 0.1 or 0.9. 3) The distribution of COCO-EE is highly uneven.

\begin{table*}[t]
  \renewcommand\arraystretch{1.05}
  \begin{center}
    \scalebox{0.95}{
    \begin{tabular}{l|c|c|cccccccc|c}
    \hline
    \multirow{2}{*}{Model} & {Unpaired} &\multirow{2}{*}{Step} &\multicolumn{8}{c|}{Quality Evaluation} & {Inference}
    \\
    & {Data} & & {B-1} & {B-2} & {B-3} & {B-4} & {R} & {C} & {S} & CLIP-Score &{Time(ms)}\\
    \hline
    Ref-Caps & {---} & {---}& 34.7  & 24.0  & 16.8  & 10.9  & 36.9  & 91.3 & 23.4  &  0.5896 & {---} \\
    \hline
    {TIger~\cite{wang2022explicit}} & \textcolor{red}{\xmark} & 3 & {31.9} & {23.9} & {18.2} & {12.4} & {40.6} & {131.8} & {30.8} & 0.6467  & 501.42\\
    {TIger-N~\cite{wang2022explicit}} & \textcolor{mygreen}{\cmark} & 3 & {33.4} & {24.1} & {17.9} & {12.2} & {39.8} & {119.8} & {28.2} &  0.6401 & 504.19\\
    \textbf{DECap} & \textcolor{mygreen}{\cmark} & 3 & \cellcolor{mygray-bg}{37.6} & \cellcolor{mygray-bg}{27.5} & \cellcolor{mygray-bg}{19.8} & \cellcolor{mygray-bg}{13.7} & \cellcolor{mygray-bg}{40.8} & \cellcolor{mygray-bg}{134.0} & \cellcolor{mygray-bg}{31.0} & \cellcolor{mygray-bg}{0.6829} & \cellcolor{mygray-bg}{214.46}\\
    \textbf{DECap} & \textcolor{mygreen}{\cmark} & 4 & \textbf{\cellcolor{mygray-bg}{38.2}} & \textbf{\cellcolor{mygray-bg}{27.9}} & \textbf{\cellcolor{mygray-bg}{20.3}} & \textbf{\cellcolor{mygray-bg}{14.1}} & \textbf{\cellcolor{mygray-bg}{41.1}} & \textbf{\cellcolor{mygray-bg}{138.2}} & \textbf{\cellcolor{mygray-bg}{31.3}} & \textbf{\cellcolor{mygray-bg}{0.6951}} & \cellcolor{mygray-bg}{282.08} \\
    \hline
    \end{tabular}%
  }
  \end{center}
  \vspace{-0.5em}
  \caption{The \emph{in-domain} evaluation of ECE models on the Flickr30K-EE test set. All models were trained on the Flickr30K-EE training set. ``Ref-Caps" denotes the initial quality of given reference captions. ``TIger-N'' denotes the TIger trained with noised unpaired data.}
  \label{tab:in_flee}%
\end{table*}%

\section{More results about DECap's generalization ability}
\label{sec:flee}

In this section, we further evaluated the generalization ability of our model with both in-domain and out-of-domain evaluation on the Flickr30K-EE~\cite{wang2022explicit}. Specifically, the Flickr30K-EE contains 108,238 training instances, 4,898 validation instances, and 4,910 testing instances, where each editing instance consists of one image and one corresponding Ref-GT caption pair. We also answered three research questions:  1) \textbf{Q1:} Does DECap perform well on the existing in-domain benchmark? 2) \textbf{Q2:} Does DECap perform well on reference captions with different noisy levels (\ie, different Levenshtein ratios)? \textbf{Q3:} Does DECap perform well on pure random reference captions? Specifically, DECap only used the unpaired data (\ie, image and GT-Cap without Ref-Cap) with diffusion step $T=6$, while the state-of-the-art TIger~\cite{wang2022explicit} was trained with the complete editing instance. For a more fair comparison, we also trained the TIger with the synthesized noised unpaired data (denoted as TIger-N). All three models used the same vocabulary sized 19,124.

\subsection{In-Domain Evaluation: Flickr30K-EE (Q1)} We compared the performance of each model on the Flickr30K-EE test set as the in-domain evaluation. And we evaluated edited captions against their single ground-truth caption.

\noindent\textbf{Results.} The in-domain evaluation results are reported in Table~\ref{tab:in_flee}. From the table, we can observe: 1) For the quality evaluation, TIger achieves its best performance using three editing steps, and our DECap achieves better results with the same step. With more editing steps, DECap can further improve the quality of captions on all metrics. \emph{It is worth noting that TIger was even trained on the in-domain Ref-GT caption pairs}.  2) TIger-N achieves limited quality improvement on the in-domain samples. 3) For the efficiency evaluation, DECap achieves significantly faster inference speed than TIger even with more editing steps. This is because that DeCap predicts edit operations and content words simultaneously but TIger needs to conduct editing by three sequential modules.

\begin{figure*}[!t]
  \centering
  \vspace{-1em}
  \includegraphics[width=1\linewidth]{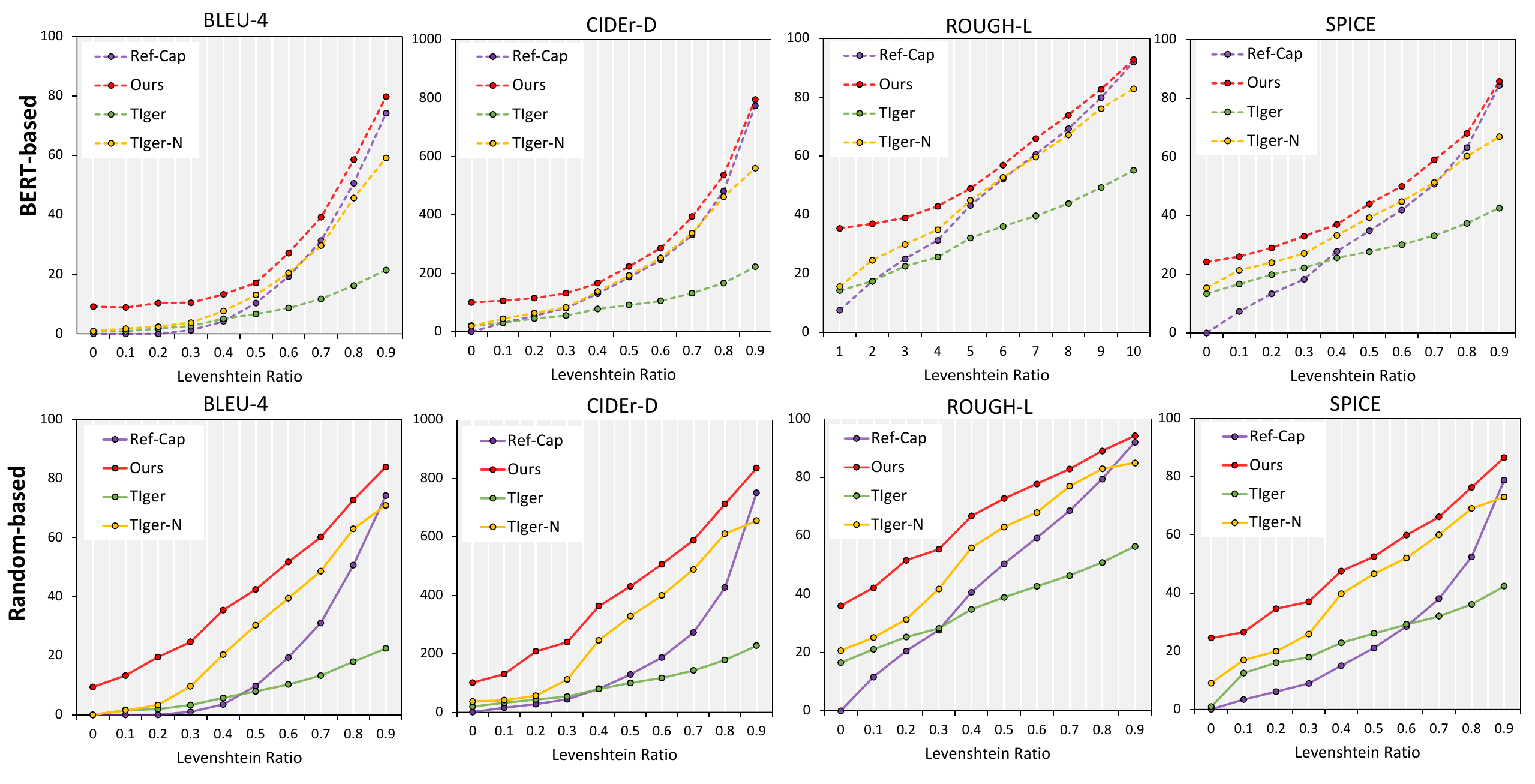}
  \caption{Performance on two kinds of out-of-domain GT-based reference captions constructed from Flickr30K-EE test set. All models were trained on the Flickr30K-EE training set. ``Ref-Caps" denotes the given reference captions, and ``TIger-N'' denotes the TIger trained with unpaired data.}
  \label{fig:out_domain2}
\end{figure*}

\subsection{OOD: GT-Based Reference Caption (Q2)}
\noindent\textbf{Setting.}  The GT-based reference captions were constructed based on the GT-Caps in the Flickr30K-EE test set. We systematically replaced words in GT-Caps with other words, resulting in the creation of various out-of-domain Ref-Caps. They varied in terms of their Levenshtein ratios, ranging from 0.9 (\ie, with only a few incorrect words) to 0.0 (\ie, where all words were wrong). Specifically, we constructed two kinds of GT-based reference captions: 1) \textbf{BERT-based}. We first replaced the GT words with the special \texttt{[MASK]} tokens and then utilized the pretrained BERT~\cite{devlin2018bert} model to predict other words different from GT words. 2) \textbf{Random-based}. We directly replaced GT words with other random words. We evaluated edited captions against their single GT-Cap.

\noindent\textbf{Results.} As shown in Figure~\ref{fig:out_domain2}, For models trained with unpaired data, our model successfully improves the quality of all kinds of the GT-based Ref captions and surpasses TIger-N. In contrast, TIger struggles when editing Ref captions with either ``minor'' or ``severe'' errors, and even degrading the captions' quality (\eg, Ref-Caps with a ratio larger than 0.4) by inadvertently removing accurate words or failing to introduce accurate details.

\begin{table*}[t]
  \renewcommand\arraystretch{1.06}
  \begin{center}
    \scalebox{0.95}{
    \begin{tabular}{l|c|c|cccccccc|c}
    \hline
    \multirow{2}{*}{Model} & {Unpaired} &\multirow{2}{*}{Step} &\multicolumn{8}{c|}{Quality Evaluation} & {Inference}\\
    & {Data} & & {B-1} & {B-2} & {B-3} & {B-4} & {R} & {C} & {S} & {CLIP-Score} &{Time(ms)}\\
    \hline
    {TIger~\cite{wang2022explicit}} & \textcolor{red}{\xmark} & 5 & {4.4} & {2.6} & {1.0} & {0.5} & {17.2} & {3.5} & {2.2} & 0.5601 & 783.00\\
    {TIger-N~\cite{wang2022explicit}} & \textcolor{mygreen}{\cmark} & 5 & 4.2 & {3.5} & {2.3} & {0.8} & {25.5} & {12.4} & {6.3} &  0.5964 & 779.50 \\
    \textbf{DECap} & \textcolor{mygreen}{\cmark} & 6 & \cellcolor{mygray-bg}{\textbf{70.1}} & \cellcolor{mygray-bg}{\textbf{47.9}} & \cellcolor{mygray-bg}{\textbf{29.8}} & \cellcolor{mygray-bg}{\textbf{17.5}} & \cellcolor{mygray-bg}{\textbf{46.7}} & \cellcolor{mygray-bg}{\textbf{45.7}} & \cellcolor{mygray-bg}{\textbf{13.3}} & \cellcolor{mygray-bg}{\textbf{0.7172}} & \cellcolor{mygray-bg}{427.00}  \\
    \hline
    \end{tabular}%
  }
  \end{center}
  \caption{Performance of ECE models on pure random (ten random words) reference captions constructed based on the Flickr30K test set. All the models were trained on the Flickr30K-EE training set. ``TIger-N'' denotes the TIger trained with noised unpaired data.}
  \label{tab:out2_flee}%
\end{table*}%

\subsection{OOD: Pure Random Reference Caption (Q3)}
\noindent\textbf{Setting.} To further evaluate the models' generalization ability without utilizing any GT captions, we constructed pure random reference captions based on the Flickr30K test set. Specifically, each editing instance consists of a single image and a Ref-Cap with ten random words. Subsequently, we evaluated the edited captions against their corresponding five GT-Caps. All results are reported in Table~\ref{tab:out2_flee}.

\noindent\textbf{Results.} From Table~\ref{tab:out2_flee}, we can observe: 1) Given the image, all the models achieve their best performance within six editing steps, DECap successfully edits the sentence with all random words into a reasonable caption. In contrast, both TIger and TIger-N face challenges in doing so. 2) In terms of efficiency metrics, DECap achieves significantly faster inference speed compared to TIger and TIger-N.

\section{Ablation Studies for the Number of Random Words}
\label{sec:num_of_words}
In this section, we run a set of ablation studies about the influence of different numbers of random words on caption generation.

\noindent\textbf{Setting.} We utilized the DECap trained with diffusion step $T=10$ on COCO training set from Sec.\textcolor{red}{4.3}, and constructed input instances consisting of an image from the COCO test set and a Ref-Cap with $n$ random words, where $n \in \{ 8,9,10,11,12\}$. The edited captions were then evaluated against their corresponding five ground-truth captions. 

 \noindent\textbf{Results.} From Table~\ref{tab:coco_num2} we can observe: 1) DECap's performance consistently improves as the number of random words increases from 8 to 10 and then starts to decline beyond 10 random words. 2) Given that the average length of ground-truth captions in COCO is around 10 words, DECap achieves its highest CIDEr-D score when editing sentences with 10 random words. While BLEU-N metrics tend to favor shorter sentences, DECap obtains the best BLEU scores with competitive CIDEr-D scores when editing sentences with 9 random words. Additionally, as the number of random words increases, DECap generates more semantic information about the image, including objects and attributes, resulting in higher SPICE scores. However, this increase in semantic content can also lead to issues like repetition and the introduction of extraneous details, referring to objects or information present in the image but not explicitly mentioned in the ground-truth captions. This can all potentially lead to a decline in the quality evaluation of the generated captions.  3) Based on these findings, we select 10 words as a balanced choice for caption generation.

\begin{table*}[t]
  \renewcommand\arraystretch{1.1}
  \begin{center}
    \scalebox{0.98}{
    \begin{tabular}{l|c|c|cccccccc}
    \hline
    {Model} & {Random Words} & {Step} & {B-1} & {B-2} & {B-3} & {B-4} & {M} & {R} & {C} & {S}
    \\
    \hline
    \textbf{} & 8 & 10 & {75.7} & {59.3} & {44.5} & {32.4} &  26.3 & {56.5} & {109.7} & {20.1}\\
    \textbf{} & 9 & 10 & \textbf{{80.2}} & \textbf{{63.3}} & \textbf{{47.9}} & \textbf{{35.5}} & 27.8 & {58.0} & {118.1} & {21.5} \\
    \textbf{DECap} & 10 & 10 & 78.0  & 61.4  & 46.4  & 34.5 & {28.6} & \textbf{{58.0}}  & \textbf{{119.0}}  & {21.9}\\
    \textbf{} & 11 &  10 & 75.9 & {58.8} & {44.3} & {32.9} & 28.9 & 57.1 & 115.7 & 22.4\\
    \textbf{} & 12 &  10 & 72.0 & {56.3} & {42.2} & {31.2} & \textbf{29.0} & 56.1 & 109.3 & \textbf{22.7}\\
    \hline
    \end{tabular}%
  }
  \end{center}
  \vspace{-0.5em}
  \caption{Performance of our model on the COCO test set with different numbers of input random words in ten editing steps.}
  \label{tab:coco_num2}%
\end{table*}%

\begin{table*}[t]
  \renewcommand\arraystretch{1.1}
  \begin{center}
    \scalebox{0.98}{
    \begin{tabular}{l|c|ccccccccc}
    \hline
    {Model} & {Distribution of Edit Types} & {B-1} & {B-2} & {B-3} & {B-4} & {M} & {R} & {C} & {S}
    \\
    \hline
    \textbf{} & $\alpha = \beta = \gamma$ & 77.4 & {60.8} & {45.8} & {34.0} & 28.6 & 57.8 & 117.4 & 21.8\\
    \textbf{DECap} & $\alpha > \beta = \gamma$ & \textbf{78.0}  & \textbf{61.4}  & \textbf{46.4}  & \textbf{34.5 }& \textbf{{28.6}} & \textbf{{58.0}}  & \textbf{{119.0}}  & \textbf{21.9}\\
    \textbf{} & $\beta = \gamma = 0$ & {{77.3}} & {{60.6}} & {{45.7}} & {{33.8}} & 25.8 & {57.8} & {116.6} & {21.7} \\
    \hline
    \end{tabular}%
  }
  \end{center}
  \vspace{-0.5em}
  \caption{Performance of our model on the COCO test set with different distributions of noising edit types.}
  \label{tab:coco_edit_type2}%
\end{table*}%

\section{Ablation Studies for the Distribution of Edit Types}
\label{sec:edit_type}
 As discussed in Sec.\textcolor{red}{3.2}, the distribution over edit types plays a crucial role in balancing different noising operations and training diverse denoising abilities. Therefore, in this section, we conduct a series of ablation experiments to examine the impact of varying distribution settings for the edit types within the edit-based noising process. Specifically, the probabilities for the noising edit operations \texttt{REPLACE}, \texttt{DELTE}, and \texttt{INSERT} is denoted as $\alpha$, $\beta$, and $\gamma$, respectively. While these probabilities are parameterized by several factors, such as the current state of the caption and the length of the ground-truth caption, we perform ablations by imposing global control over these probabilities. For instance, we explore settings where $\alpha$$=$$\beta$$=$$\gamma$, $\alpha \textgreater \beta$$=$$\gamma$ and $\beta$$=$$\gamma$$=$$0$.

 \noindent\textbf{Setting.} We train the DECap on the COCO training set with different distributions of edit types with the same diffusion set $T=10$. During testing, we constructed input instances consisting of a single image from the COCO test set and a Ref-Cap with ten random words. The edited captions were then evaluated against their corresponding five GT-Caps.

 \noindent\textbf{Results.} From Table~\ref{tab:coco_edit_type2}, we can observe: 1) In comparison to the even distribution of edit types, where $\alpha$$=$$\beta$$=$$\gamma$, DECap demonstrates improved performance when we emphasize the denoising ability of the replacement operation with the distribution $\alpha \textgreater \beta$$=$$\gamma$. This suggests that the replacement operation is more flexible and efficient in correcting words than the sequence operation of first deletion and then insertion. 2) When we trained DECap with an exclusive focus on the replacement operation and omitted the deletion and insertion abilities, setting $\beta$$=$$\gamma$$=$$0$, there is a noticeable decline in the quality of generated captions. This indicates that DECap's ability to adjust caption length by adding more description or removing repetitions is compromised. 3) These results suggest that the distribution with $\alpha \textgreater \beta$$=$$\gamma$ could be a sensible choice for caption generation. Importantly, our method allows for flexible adaptation, enabling us to set different edit type distributions tailored to specific tasks or requirements. 
 

\section{Potential Negative Societal Impacts}
\label{sec:social}
Our proposed ECE model may face the same potential ethical concerns as other existing ECE or image captioning works, such as suffering from severe bias issues (\eg, gender bias~\cite{hendricks2018women}). Additionally, our method may also be maliciously utilized by using some improper control words, such as sensitive attributes. Apart from these general issues that already exist in the ECE or image captioning tasks, our paper has no additional ethical issues.

%
%

\end{document}